\csname docmute\endcsname
\pdfoutput=1 

\newif\ifissvjour
\newif\ifiselsarticle
\newif\ifisreport

\issvjourfalse
\iselsarticletrue
\isreportfalse

\ifissvjour
	\documentclass[twocolumn]{svjour3} %
\fi
\ifiselsarticle	
	\documentclass{elsarticle}
\fi
\ifisreport
	\documentclass[11pt,a4paper]{report}
\fi

\usepackage{graphicx,rotating} 
\usepackage{tikz}
\usepackage{pstricks}
\usepackage[hang, small, bf, margin=20pt, tableposition=top]{caption} 
\usepackage{subcaption}
\usepackage{color} 
\usepackage{eso-pic} 
\usepackage{everyshi} 
\usepackage{epstopdf} 
\usepackage{float} 

\usepackage{hyperref}

\usepackage{amsmath}
\usepackage{amssymb}

\usepackage{pdflscape}
\usepackage{ifdraft}
\ifissvjour
\usepackage[switch]{lineno}
\fi
\ifiselsarticle
\usepackage[left]{lineno}
\fi

\begin{document}
\ifdraft{\linenumbers}{}

\ifiselsarticle
\begin{frontmatter}
	\title{Multi-scale Discriminant Saliency with Wavelet-based Hidden Markov Tree Modelling}
	\author[unmc,unuk]{Anh Cat \textsc{Le Ngo}}
	\author[edwu]{Kenneth Li-Minn \textsc{Ang}}
	\author[swuc]{Jasmine Kah-Phooi \textsc{Seng}} 
	\author[unuk]{Guoping \textsc{Qiu}}
	
	\address[unmc]{School of Engineering, The University of Nottigham, Malaysia Campus, 43500 Jalan Broga, Semenyih Selangor, Malaysia}
	\address[unuk]{School of Computer Science, The University of Nottingham, Jubilee Campus, Nottingham NG8 1BB  United Kingdom}
	\address[edwu]{Centre for Communications Engineering Research, Edith Cowan University, 270 Joondalup Dr  Joondalup WA 6027, Australia}
	\address[swuc]{Department of Computer Science \& Networked System, Sunway University, Jalan Universiti  Bandar Sunway, 46150 Petaling Jaya, Selangor}	
\fi

\ifissvjour 	
	\title{Multi-scale Discriminant Saliency}
	\subtitle{Decision-Theoretic Saliency Method with Wavelet-based Hidden Markov Tree Modelling}
	\author{Anh Cat Le Ngo \and Kenneth Li-Minn Ang \\ Guoping Qiu  \and Jasmine Kah-Phooi Seng}
	\institute{
	Anh Cat Le Ngo \at School of Engineering, The University of Nottingham, Malaysia Campus \\ \email{keyx9lna@nottingham.edu.my} 
	\and 
	Kenneth Li-Minn Ang \at Centre for Communications Engineering Research, Edith Cowan University
	\and 
	Guoping Qiu \at School of Computer Science, The University of Nottingham, UK Campus
	\and
	Jasmine Kah-Phooi Seng \at Department of Computer Science \& Networked System, Sunway University 
	}
	\maketitle
\fi	
\ifiselsarticle
\begin{linenomath*}
\fi
\ifiselsarticle
\begin{abstract}
\fi
\ifissvjour
\begin{abstract}
\fi	
	Bottom-up saliency, an early stage of human visual attention, can be considered as a binary classification problem between centre and surround classes. Discriminant power of features for the classification is measured as mutual information between distributions of image features and corresponding classes . As the estimated discrepancy very much depends on considered scale level, this \ifisreport\else paper \fi proposes computing discriminant power in multi-scale structure with employment of discrete wavelet features and Hidden Markov Tree (HMT). With wavelet coefficients and Hidden Markov Tree parameters, quad-tree like label structures are constructed and utilized in maximum a posterior probability (MAP) of hidden class variables at corresponding dyadic sub-squares. A saliency value for each square at each scale level is computed with discriminant power principle. Finally, across multiple scales is integrated the final saliency map by an information maximization rule. Both standard quantitative tools such as NSS, LCC, AUC and qualitative assessments are used for evaluating the proposed multi-scale discriminant saliency (MDIS) against the well-know information based approach AIM on its accompanied image collection with eye-tracking data. Simulation results are presented and analysed to verify the validity of MDIS as well as point out its limitation for further research direction.
\ifiselsarticle
\end{abstract}
\fi
\ifissvjour
\end{abstract}
\fi	
\ifiselsarticle
\end{linenomath*}
\fi
\ifiselsarticle
\end{frontmatter}
\fi

\section{Visual Attention - Computational Approach }
\label{sec:intro_journal2_2013}
Visual attention is a psychological phenomenon in which human visual systems are optimized for capturing scenic information. Robustness and efficiency of biological devices, the eyes and their control systems, visual paths in the brain have amazed scientists and engineers for centuries. From Neisser \cite{neisser1967} to Marr \cite{marr1976}, researchers have put intensive effort in discovering attention principles and engineering artificial systems with equivalent capability. For decades, this research field has been dominated by visual attention principles, proposing existence of saliency maps for attention guidance. The idea is further promoted in Feature Integration Theory (FIT) \cite{treisman1980} which elaborates computational principles of saliency map generation with centre-surround operators and basic image features such as intensity, orientation and colour. Then, Itti et al. \cite{itti1998} implemented and released the first complete computer algorithms of FIT theory \footnote{http://ilab.usc.edu/toolkit/}.

Feature Integration Theory is widely accepted as principles behind visual attention partly due to its utilization of basic image features such as colour, intensity, and orientation. Moreover, this hypothesis is supported by several evidences from psychological experiments. However, it only defines theoretical aspects of saliency maps and visual attention, but does not investigate how such principles would be implemented algorithmically. This lack of implementation details leaves the research field open for many later saliency algorithms \cite{itti1998,gao2004,sun2003,harel2007}, etc. Saliency might be computed as a linear contrast between features of central and surrounding environments across multiple scales by centre-surround operators. Saliency is also modelled as phase difference in Fourier Transform Domain \cite{hou2007}, or its value depends on statistical modelling of the local feature distribution \cite{sun2003}. Though many approaches are mentioned in a long and rich literature of visual saliency, only a few are built on a solid hypothesis or linked to other well-established computational theories. Among these approaches, Neil Bruce's work \cite{bruce2006} has nicely established a bridge between visual saliency and information theories. It puts a first step for bridging two alien fields; moreover, visual attention for first time can be viewed as information system. Then, information based visual saliency has continuously been investigated and developed in several works \cite{lengo2012,lengo2010,qiu2007,gao2007-a}. The distinguishing points between these works are computational approaches for retrieving information from features. Then, the process attracts much interest from research community due to its grand challenges in estimating information of high-dimensional data like 2-D image patches. It usually runs into computational problems which can not be efficiently solved due to the curse of dimensionality; moreover, central and surrounding contexts are usually defined in ad-hoc manners without much theoretical supports. To tackle these problems, Dashan Gao et al. \cite{gao2007-a} has simplified the information extraction step as a binary classification process between centre and surround contexts. Then the discriminant power or mutual information between features and classes are estimated as saliency values for each location. 

Spatial features have large influence on saliency values; however, scale-space features do have decisive roles in visual saliency computation since centre or surround environments are simply processing windows with different sizes. In signal processing, scale-space and spectral space are two sides of a coin; therefore, there is a strong relation between scale-frequency-saliency in visual attention problem. Several researchers \cite{baddeley2006,reinagel1999,parkhurst2002,tatler2005} have outlined that fixated regions have high spatial contrast or showed that high-frequency edges allow stronger discrimination fixated over non-fixated points. In brief, they all come up with one conclusion: increment in predictability at high-frequency features. Although these studies emphasizes a greater visual attraction to high frequencies (edges, ridges, other structures of images), there are other works focusing on medium frequency. 
 Perhaps, that attention system may involve different range of frequencies for optimal eye-movements. In other words, the diversity of attention in spectral spaces needs necessary utilization of scale-space theory. 

Though multi-scale nature has been emphasized as the implicit part of human visual attention, it is often ignored in several saliency algorithms. For example, DIS approach \cite{gao2009} considers only one fixed-size window; hence, it may lead to inconsideration of significant attentive features in a scene. Therefore, DIS approach needs constituting under the multiscale framework to form multi-scale discriminant saliency (MDIS) approach. This is the main motivation as well as contribution of this \ifisreport\else paper \fi which are organized as follows. Section \ref{sec:dis} reviews principles behind DIS \cite{gao2007-a} and focuses on its important assumption and limitation. 
After that, MDIS approach is carefully elaborated in section \ref{sec:mdis} with several relevant contents such as multiple dyadic windows for binary classification problem in subsection \ref{subsec:mcw}, multi-scale statistical modelling of wavelet coefficients and learning of parameters in sub-sections \ref{subsec:msm}, \ref{subsec:msl}, maximum likelihood (MLL) and maximum a posterior probability (MAP) computation of dyadic sub-squares in subsections \ref{subsec:mlc}, \ref{subsec:map}. Then, all MDIS steps are combined for final saliency map generation in subsection \ref{subsec:mdis}. Quantitative and qualitative analysis of the proposal with different modes are discussed in section \ref{sec:exp}; moreover, comparisons of the proposed MDIS and the well-known information-based saliency method AIM \cite{bruce2006} simulation data are presented with several interesting conclusions. Finally, main contributions of this paper as well as further research direction are stated in the conclusion section \ref{sec:conclusion_journal2_2013}.
\section{Visual Attention - Discriminant Saliency}
\label{sec:dis}
Saliency mechanism plays a key role in perceptual organization, recently several attempts are made to generalize principles for visual saliency. From the decision theoretic point of view, saliency is regarded as power for distinguishing salient and non-salient classes; moreover, it combines the classical centre-surround hypothesis with a derived optimal saliency architecture. In other word, saliency of each image location is identified by discriminant power of a feature set with respect to the binary classification problem between centre and surround classes. Based on decision theory, the discriminant detector can work with variety of stimulus modalities, including intensity, colour, orientation and motion. Moreover, various psychophysical properties for both static and motion stimuli are shown to be accurately satisfied quantitatively by DIS saliency maps. 
\ifisreport
\else
With regards to the theory, centre and surround classes are respectively defined as two following hypotheses.
\fi
\begin{itemize}
	\item Interest hypothesis: observations within a central neighborhood \(W_{l}^{1}\) of visual fields location \(l\).
	\item Null hypothesis: observations within a surrounding window \(W_{l}^{0}\) of the above central region.
\end{itemize}

Within DIS, feature responses within the windows are randomly drawn from the predefined sets of features \(X\). Since there are many possible combinations and orders of how such responses are assembled, the observations of features can be considered as a random process of dimension \(d\). \(X(l)=(X_{1}(l),\ldots,X_{d}(l))\). Each observation is drawn conditionally on the states of hidden variable \(Y(l)\), which is either centre or surround state. Feature vectors \(\mathbf{x(j)}\) such that \(j\in W_{l}^{c},c\in\{0,1\}\}\) are drawn from classes \(c\) according to conditional densities \(P_{X(l)}\vert_{Y(l)}(x|c)\) where \(Y(l)=0\) for surround or \(Y(l)=1\) for centre. The saliency \(S\) at location \(l\), \(S(l)\), is equal to the discriminant power of X for classifying observed feature vectors, which can be quantified by the mutual information between features, X, and class labels, Y. 
\ifisreport
\else
\begin{alignat}{2}
S(l) &=&& \sum_{c}\int p_{X,Y}(x,c)log\frac{p_{X,Y}(x,c)}{p_{X}(x)p_{y}(c)}dx\\
	&=&& \frac{1}{\vert W_{l}|}\sum_{j\in W_{l}}\left[H(Y)+\sum_{c=0}^{1}P_{Y\vert X}(c\vert x_j)log\, P_{Y\vert X}(c\vert x_j)\right] 
 \label{eq:dis}
\end{alignat}
\fi
Given a location \(l\), there are corresponding centres \(W_{l}^{1}\) and surround \(W_{l}^{0}\) windows along with a set of associated feature responses \(x(j), j\in W_{l}=W_{l}^{0}\cup W_{l}^{1}\). 
\section{Multiscale Discriminant Saliency}
\label{sec:mdis}
Expansion from a fixed window-size to multi-scale processing is commonly desired in development of computer vision algorithms. In long literature of computer vision research fields, there are many multi-scale processing models which can be used as references for developing a so called Multi-scale Discriminant Saliency (MDIS). Any selected model has to adapt binary classification in multi-scale stages. Put differently, it requires efficient classification of imaging data into a class at a particular scale with prior knowledge from other scales. With respect to these requirements, a multi-scale image segmentation framework should be a great starting point for MDIS since DIS can be considered as simplified binary image segmentation with only two classes. However, the binary classification is only an intermediate step to measure discriminant power of centre-surrounding features, and accuracy of segmentation results does not really matter in this case.

Typical algorithms employ a rigid classification window in a vague hope that all pixels in region of interest (ROI) belong to the same class. Obviously, DIS has similar problems of choosing suitable window sizes as well. Clearly, the size is crucial to balance between its reliability and accuracy. A large window usually provides rich statistical information and enhance reliability of the algorithm. However, it also risks including heterogeneous elements in the window and eventually loses segmentation accuracy. Therefore, appropriate window sizes are equivalently vital in avoidance of local maxima in calculation discriminant power. If window sizes are too large or too small, MDIS risks losing useful discriminative features or being too susceptible to noise. In brief, sampling rates, and consequently number of data in a window, directly affect on performance of binary classification / segmentation and eventually computation of discriminating powers.

\subsection{Multi-scale Classification Windows}
\label{subsec:mcw}
Multi-scale segmentation employs multi-scale classification windows on images, then combines responses across scales. MDIS can adapt similar approach to classify image features into either centre or surrounding class. Though window sizes are preferably chosen arbitrarily, dyadic squares ( or blocks ) are implemented in MDIS because of its compactness and efficiency. Let's assume an initial square image \(s\) with \(2^J\)x\(2^J\) of \(n:=2^{2J}\) pixels, the dyadic square structures can be generated by recursively dividing \(x\) into four square sub-images equally, left-side of the figure \ref{fig:waveletstr}. As a result, it becomes the popular quad-tree structure, commonly employed in computer vision and image processing problems. In this tree structure, each node is related to a direct above parent node while it plays a role of parental nodes itself for four direct below nodes \ref{fig:waveletstr}. Each quad-tree node is equivalent with a dyadic square, and is denoted as a tree-node in scale j by \(d_{i}^{j}\) whereof \(i\) is a spatial index of a dyadic square node. Given a random field image \(X\), the dyadic squares are also random fields, formulated as \(D_{i}^{j}\) mathematically. In following sections, we sometime use \(D_{i}\) (dropping scale factor \(j\)) as general randomly-generated dyadic square regardless of scales.
\begin{figure}[!htbp]%
\centering
\includegraphics[width=\columnwidth]{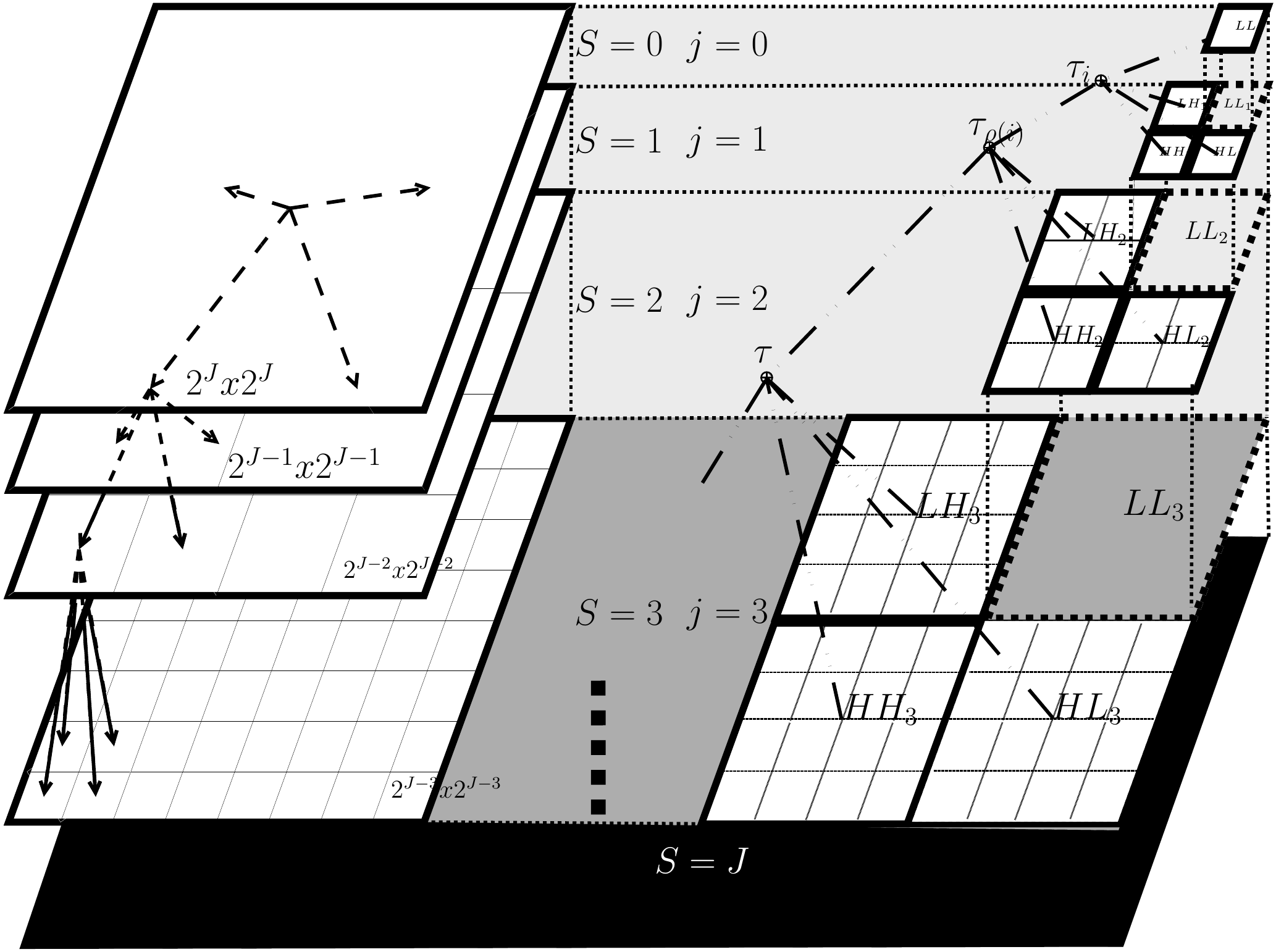}
\caption{Quad-Tree and Dyadic Wavelet Structures}
\label{fig:waveletstr}
\end{figure}
Using these dyadic squares as classification windows, we can classify each node \(d_i\) as either centre or surround by estimating its maximum a posterior probability (MAP). Then, mutual information between features and corresponding labels is averages of MAP across all classes. Mutual information is similar to the core concept of discriminant power for central features against surrounding ones at each location \cite{gao2004}. However, deployment of quad-tree structures makes the estimation of mutual information possible for many scales. In order to compute the discriminant power, multiple probability distribution functions (PDF) need to be learned through wavelet-based statistical models.

\subsection{Multi-scale Statistical Model}
\label{subsec:msm}
Hidden Markov Tree (HMT) captures the main statistical characteristics in wavelet coefficients of real-world images. On one hand, parameters originally have to be learnt for each data point in a raw model of HMT; however, such training makes it unwieldy for handling images with large amount of data. On the other hand, arbitrarily specified parameters would help to avoid a time-consuming learning stage which is even impossible in some cases, but would risk of over-fitting the model. The above issues may render HMT inappropriate for applications with rapid processing requirement but without sufficient priori information. Therefore, several derivatives of algorithms, based on HMT, are studied in this section in order to make HMT more computationally feasible as well as find out their advantages and limitations.

Marginal distributions of wavelet coefficients \(w_i\) come directly from sparseness of wavelet transform in modelling real-word images: a minority of large coefficients and a majority of small coefficients. That distribution is efficiently captured by GMM with wavelet coefficients \(w_i\) and observed hidden state variables or class labels  \(S_i \in {S,L}\). A state value \(S_i\) decides which mixture generates coefficients \(w_i\). 
\begin{align}
g(x;\mu,\sigma^2) := \frac{1}{\sqrt{2\pi}\sigma} exp{-\frac{(x-\mu)^2}{2*\sigma^2}}
\end{align}
Lets denote the Gaussian Probability Distribution Function (PDF) of small-variance state \(S_i=S\) is as follows.
\begin{align}
f(w_i\vert S_i=S) = g(w_i;0,\sigma_{S;i}^2) 
\end{align}
While state \(S_i = L \) has zero-mean, large-variance Gaussian
\begin{align}
f(w_i \vert S_i = L) = g(w_i;0,\sigma^2_{L;i}) 
\end{align}
where \(\sigma^2_L \ge \sigma^2_S\). By those mixture models, we can write the marginal PDF \(f(w_i)\) as a convex combination of the conditional densities.
\begin{align}
f(w_i) = p^S_{i}g(w_i;0,\sigma_{S;i}^2) + p^L_i g(w_i;0,\sigma_{L;i}^2)
\end{align}
where \(p^S_i+p_i^L=1\) since \(p_{S_i}=[p^S_i p^L_i]\) are mass probability of states. In statistical interpretation, \(p_{S_i}\) is how likely wavelet coefficients \(w_i\) are small or large. 

HMT captures inter-scale dependencies of probabilistic tree connecting hidden state variables of a wavelet coefficient and its four children. Therefore, dependency graphs have similar quad-tree topology as wavelet decomposition, and it includes state-to-state links between parent and child coefficients, mathematically modelled by persistency probabilities and novelty probabilities.
\begin{align}
A_i = \left[ \begin{array}{cc}
p_i^{S \rightarrow S}  & p_i^{S \rightarrow L} \\
p_i^{L \rightarrow S}  & p_i^{L \rightarrow L}
\end{array} \right]
\end{align}
where \(p_i^{S \rightarrow S} + p_i^{S \rightarrow L} =1 \) and \(p_i^{L \rightarrow S} + p_i^{L \rightarrow L} = 1\). Persistency probabilities are lying on the main diagonal axis of the above array \(p_i^{S \rightarrow S}\),\(p_i^{L \rightarrow L}\), since they represent how likely states are kept in parent and child links. On the sub diagonal axis are novelty of probabilities, which causes different states between parent and child nodes. 

In summary, trained HMT models can be specified in terms of (i) GMM mixture variances \(\sigma_{S;i}^2\) and \(\sigma_{L;i}^2\) (ii) the state transition matrix \(A_i\) and (iii) probability mass function \(p_1\) at the coarsest level. Grouping these parameters into a model \({\cal M}_c\), we can define trained HMT model as follows
\begin{alignat}{2}
{\cal M}_c &=&& \{ p_1,A_2,\ldots,A_J;\sigma_{S_i;k,j,b}\} \\
\mbox{where}&&& \nonumber \\ 
j 		   &=&&  1,\ldots,J \\ 
S_i		   &=&&  \{L,S\} \\ 
k 		   &\in&& \mathbb{Z}^2 \\ 
\forall b  &\in&& {\cal B} 
\end{alignat}
where \(k\) is wavelet coefficients index at scale \(j\) while \(b\) represents wavelet sub-bands. 
\begin{align}
{\cal B} = \{HL,LH,HH\}
\end{align}
The parametric model \({\cal M}_c\) can be learned from the joint $PDF$ \(f(\mathbf{w\vert {\cal M}_c})\) of the wavelet coefficients in each of three sub-bands \cite{crouse1998}. Generally, each node of wavelet coefficients has its own model with different parameters. However, it overcomplicates the model with too many parameters; for examples, \(n\) wavelet coefficients are required to be fitted on $4n$ parameter models which is an impossible task. Therefore, to reduce the complexity, HMT is assumed to use same parameters for every node at the same scale of the wavelet transform regardless of spatial \(k\) and oriental \(b\) indexes.
\begin{alignat}{2}
\sigma^2_{S;b,j,k} &=&& \sigma^2_{S;j} \\
\sigma^2_{L;b,j,k} &=&& \sigma^2_{L;j} \\
 A{b,j,k} &=&& A_j \\
 k \in \mathbb{Z}^2 &,&& \forall b \in {\cal B}
\end{alignat}
The assumption is called \emph{tying within scale} \cite{crouse1998} and it prevents over-complex and infeasible HMT model in exchange of less general model, which is mathematically formulated as follows.
\begin{align}
{\cal M}_c = \{ p_1,A_2,\ldots,A_J;\sigma_{S_i;j},(j=1,\ldots,J,S_i={L,S}) \}
\label{eq:thmt_model}
\end{align}
Lets name the \({\cal M}_c\) in equation \ref{eq:thmt_model} as trained HMT (THMT). While the \emph{typing} trick has significantly simplified the learning process for HMT parameters. In fact, the model only needs training on an input image. However, it possible to further simplify the approach if the image category is known in advance. Using many images with similar contexts, we can train HMT offline for meta-parameters , which are later fixed in an HMT model. It yields a general HMT model for that class of images with each member in the class being treated statistically equivalent \cite{romberg2001}. Fortunately, Romberg \cite{romberg2001} have studied similar models for the class of natural images and published their universal parameters \({\cal M}_c\) obtained by (jointly) fitting lines to the HMT parameters of natural images
\ifisreport
\else
.
\fi

The variance and persistence decays are measured by fitting a line to the log of the variance versus scale for each state \cite{romberg2001}, and it is only started at scale \(j=4\) with transition state probability at \(j=5\). This choice of scale ensures enough data for an accurate estimate of decays; moreover, Romberg .et .al \cite{romberg2001} find out that the decays are very similar for many of the natural images. Therefore, it is reasonable to fix the HMT model with meta-parameters which are learnt from natural images. This approach of modelling HMT is called Universal Hidden Markov Tree (UHMT) in this paper. Though accuracy of this UHMT model is clearly lost by treating all different images statistically equivalent, its assumption can totally eliminate the need for training and save tremendous computational workload, and make real-time HMT possible. Experiments with UHMT mode are mentioned in the section \ref{sec:exp} where both accuracy and efficiency of UHMT are evaluated. 

As UHMT approach eliminates training stages of THMT to decrease computational requirement for real-time applications, sometimes it is necessary to have better HMT in modelling image. Though \emph{tying} THMT assumes the same parameters for wavelet sub-bands or coefficients at different orientations, its underlying learning treats each brand independently. However, experiments by Simoncelli and Portilla \cite{simoncelli1998} demonstrate importance of cross-correlation between wavelet sub-bands with different orientations at the same scale for modelling texture image. Moreover, textural features are pretty common in natural images as well; therefore, capturing this dependency would improve accuracy of HMT models. Since THMT treats coefficients of sub-bands independently, it obviously ignores the cross-orientation correlation. To enhance the capacity of THMT, Do and Vetterli \cite{do2002} propose grouping of coefficients at the same location and scale into a vector, then carry out HMT modelling in a multidimensional manner. In this paper, we call such approach as Vectorized HMT (VHMT). Let's denote a vector after the grouping of wavelet coefficients at location \(k\), scale \(j\) in three different orientations vertical \((V)\), horizontal \((H)\), and diagonal \((D)\) as follows.
\begin{align}
\mathbf{w}_{j,k}=(w_{j,k}^{H} w_{j,k}^{V} w_{j,k}^{D})^T
\end{align}
As multidimensional groups of wavelet coefficients \(\mathbf{w}_{j,k}\) are concerned, their distribution need formulating as zero-mean multivariate Gaussian density with covariance matrix \(C\) as follows
\begin{align}
g(\mathbf{w};C) = \frac{1}{\sqrt{(2\pi)^n \vert det(C) \vert}} \mbox{exp}(-\mathbf{w}^T C^{-1} \mathbf{w}) 
\end{align}
where n is the number of dimensions, or orientations \(n=3\) in this case. Except from a multivariate probability density, VHMT follows similar steps as other HMTs do. Its marginal distribution is formulated as Gaussian Mixture Models, i.e.
\begin{align}
f_j(\mathbf{w}) = p_j^L g(\mathbf{w};C_j^L) + p_j^S g(\mathbf{w};C_j^S)
\end{align}
Moreover, its statistical inter-scale dependency is modelled through the parent-child relationship with a quad-tree structure linking a parent with its four children at the next level in the same location. A small difference here is that only one tree is utilized instead of three trees since wavelet coefficients have been grouped and modelled simultaneously. Hence, an image can be modelled by VHMT with a set of parameters.
\begin{align}
	\theta=\{ \mathbf{p_1},A_2,\ldots,A_J;C_j^{S_i},(j=1,\ldots,J,S_i={L,S})\}
\end{align}
As only one quad-tree is built for modelling a vector of wavelet coefficients, the hidden states are as well "tied up". It means the same hidden state is assigned regardless of orientations; in other words, VHMT is orientation-invariant. VHMT captures dependencies across orientations via a covariance matrix \(C\) of a multivariate Gaussian density. Diagonal elements of the matrix are variances of each orientation meanwhile non-diagonal elements represent covariance of wavelet coefficients across sub-bands. It justifies the VHMT model for textural features because their wavelet coefficients have high possibility of being significant at all orientations in edge regions whereas they are likely small at any directions in smooth regions. 
\subsection{Multi-scale Statistical Learning}
\label{subsec:msl}

The complete joint pixel PDF is typically overcomplicated and difficult to model due to their high-dimensional nature. Unavailability of simple distribution model in practice motivates statistical modelling of transform-domain which is often less complex and easier to be estimated. Obviously, joint pixel PDF could be well approximated as marginal PDF of wavelet coefficients. Since the transform well-characterizes semantic singularity of natural images, it provides a suitable transform-domain for modelling statistical property of singularity-rich images. 

The singularity characterization along scales makes the wavelet domain well-suited for modelling natural images. In fact, statistical models of wavelet coefficients have quite comprehensive literature; however, we only concentrate on the Hidden Markov Tree of Crouse, Nowak and Baraniuk \cite{crouse1998}. In consideration of both marginal and joint statistics of wavelet coefficient, the HMT model introduces a hidden state variable of either "large" or "small" to each quad-tree node at a particular scale. Then, the marginal density of wavelet coefficients is modelled as a two-states Gaussian mixture in which a "large" or "small" refers to characteristics of Gaussian distribution's variance values. The mixtures closely match marginal statistics of natural images \cite{pesquet1996},\cite{abramovich1998}, \cite{chipman1997}. With the HMT, persistence of large or small coefficients is captured across scales using Markov-1 chain. It models dependencies of hidden states across scale in a tree structure, parallel to those of wavelet coefficients and dyadic squares. With parameters of GMM and Markov State Transition in vector \(\cal{M}\), the HMT model is able to approximate overall joint PDF of wavelet coefficients \(\mathbf{W}\) by a high-dimensional but highly structured Gaussian mixture models \(f(\mathbf{w} \vert \cal{M})\). 

Highly structural nature of wavelet coefficients allows efficient implementation of HMT-based processing. The parameters of a HMT model  \(\cal{M}\) can be learned through an iterative expectation and maximization (EM) algorithm with cost \({\cal O}(n)\) per iteration \cite{crouse1998} in (T/V)HMT or predefined for a particular image category \cite{romberg2001} in UHMT. After the parameters \(\cal{M}\) are estimated by the EM iteration, we need to compute correspondent statistical characteristics given DWT coefficients \(\mathbf{\tilde{w}}\) of an image \(\mathbf{\tilde{x}}\) and a set of HMT parameters \({\cal M}\). It is a realization of the HMT model in which computation of the likelihood \(f(\mathbf{\tilde{\omega}\vert{\cal{M}})}\) requires only a simple \({\cal O}(n)\) up-sweep through the HMT tree from leaves to root \cite{crouse1998}. 
This model opens a prospect of a simple multi-scale image classification algorithm. Supposed centre and surround classes are denoted as \(c\in \{1,0\}\), we have specified or trained HMT trees for each class with parameters \({\cal M}_{c}\). Then the above likelihood calculation is deployed on each node of the HMT quad-tree given the wavelet transform of an image \(\mathbf{\tilde{x}}\). For each node of the tree, HMT yields likelihood \(f(\tilde{\mathbf{d_{i}}|{\cal M}_{c}),}c\in\{1,0\}\) for each dyadic block \(d_{i}\). With these multi-scale likelihoods, we can easily choose the most suitable class \(c\) for a dyadic sub-square \(\mathbf{\tilde{d_i}}\) as follows.
\begin{align}
\hat{c}_{i}^{ML}:=argmax_{c \in \{1,0\}}f(\mathbf{\tilde{d}_{i}}|{\cal M}_{c})
\end{align}

\subsection{Multiscale Likelihood Computation}
\label{subsec:mlc}
To obtain likelihood under a sub-tree \({\cal T}_i\) of wavelet coefficients rooted at \(w_i\), we have deployed wavelet HMT trees and learn parameters \(\Theta\) for multiple levels \cite{crouse1998}. The conditional likelihood \(\beta_i(m):=f({\cal T}_i|S_i = m,\Theta)\) can be retrieved by sweeping up to node \(i\) (see \cite{crouse1998}); then, likelihood of a coefficient in \({\cal T}_i\) can be computed as follows.
\begin{align}
f({\cal T} \vert \Theta) = \sum_{m=S,L} \beta_i(c)p(S_i = c \vert \Theta)
\label{eq:llh}
\end{align}
with \(p(S_i=c\vert\Theta)\) state probabilities can be predefined or obtained during training \cite{romberg2001}. 

Due to similarity between sub-trees of wavelet coefficients and dyadic squares, it is obvious that pixels of each square block \(\mathbf{d_i}\) are well represented by three sub-bands or sub-trees \(\{ {\cal T}_i^{LH}, {\cal T}_i^{HL}, {\cal T}_i^{HH}\}\), whereof all likelihood are independently calculated by the equation \ref{eq:llh} in their corresponding trees . Independent estimation of three bands is an appropriate computation step due to an assumption that correlation between feature channels would not affect discriminant powers Gao \cite{gao2009}. Furthermore, DWT is known as de-correlation tools as well, and decorrelated signals are linearly independent from each other. Hence, the likelihood of a dyadic square is formulated as product of three independent likelihoods of wavelet sub-bands at each scale.
\begin{align}
	f(d_i \vert {\cal M} ) = f({\cal T}_{i}^{LH}\vert \Theta^{LH})f({\cal T}_{i}^{HL}\vert \Theta^{HL})f({\cal T}_{i}^{HH}\vert \Theta^{HH})
\label{eq:blh_thmt}
\end{align}
Noteworthy that, assumption of independent sub-bands is only necessary in the (U/T) HMT model while VHMT has grouped all coefficients into a single vector. Therefore, it only needs one quad-tree representation of multivariate coefficients, $\vec{{\cal T}}$,  and likelihood of dyadic squares under each tree node is formulated as follows.
\begin{align}
	f(d_i \vert {\cal M} ) = f(\mathbf{{\cal T}_{i}}\vert \Theta)
\label{eq:blh_thmt_vhmt}
\end{align}
The above simple formulation of likelihood is usually employed in block-by-block or "raw" classification since it does not exploit any possible relationship at different scales. Moreover, classification decisions between classes (centre and surround) are lack of inheritance across dyadic scales because a process of likelihood estimation at each scale is isolated from processes at other levels. Therefore, a better classifier can be achieved by integrating prior knowledge of other scales or at least the direct coarser scale.
\subsection{Multi-scale Maximum a Posterior}
\label{subsec:map}
In the previous section, only "raw" binary classification between two states has been realized under the wavelet HMT model \cite{crouse1998}. Given a prior knowledge from other scales, a better binary classification solution for DIS and MDIS, the equation \ref{eq:dis}, needs a posterior probability \(p(c_{i}^{j}|d^j)\) whereof \(c_{i}^{j}\) and \(d^j = d_{i}^{j}\) are respectively class labels and features of an image at a dyadic scale \(j\) and location \(i\). 

In order to estimate the MAP \(p(c_{i}^{j}|d^j)\), we need to employ Bayesian approach for capturing dependencies between dyadic squares at different scales. Though many approximation techniques \cite{cheng1997},\cite{cheng1998},\cite{Li2000},\cite{bouman1994} are derived for a practical computation of MAP, the Hidden Markov Tree (HMT) by Choi \cite{choi2001} is proven to be a feasible solution. Choi \cite{choi2001} introduces hidden label tree modelling instead of joint probability estimation in high-dimensional data of dyadic squares. Due to strong correlation of a square under inspection with its parents and their neighbours, class labels for these adjacent squares would affect the class decision at the considered square. For example, if the parent square belongs to a certain class and so do their neighbours, the child square most likely belongs to the same class as well. Therefore, the parent-child relation should be modelled by a general probabilistic graph \cite{crouse1998}. However, the complexity exponentially increases with number of neighbouring nodes. Choi \cite{choi1998} proposes an alternative simpler solution, based on context-based Bayesian approach. For the sake of simplicity, causal contexts are only defined by states of the direct parent node and its 8 intermediate neighbours. Let's denote the context for \(D_i\) as \(\mathbf{v_i} \equiv [v_{i,0},v_{i,1},\ldots,v_{i,8}]\) where \(v_{i,0}\) refers to the neighbours of the direct parent node and their neighbours. The triple \(\mathbf{v_i} \rightarrow C_i \rightarrow \mathbf{D_i}\) forms a Markov-1 chain, relating prior context \(\mathbf{v_i}\) and node features \(\mathbf{D_i}\) to classification decisions \(C_i\). Moreover, class labels of prior contexts \(\mathbf{v_i}\) are chosen as discrete values as it simplifies the modelling considerably. Given that prior context, independence can be assumed for label classification at each node; therefore, it is allowed to write.
\begin{align}
p(\mathbf{c^j} \vert \mathbf{v^j}) = \prod_{i}{p(c_{i}^{j}) \vert \mathbf{v_{i}^{j}}}
\end{align}
The property of Markov-1 chain assumes that \(\mathbf{D_i}\) is independent from \(\mathbf{V_i}\) given \(C_i\); therefore, the posterior probability of classifying \(\mathbf{c^j}\) given \(\mathbf{d^j,v^j}\) is written as follows.
\begin{align}
p(\mathbf{c^j \vert d^j,v^j}) = \frac{f(\mathbf{d^j \vert c^j})p(\mathbf{c^j \vert v^j})}{f(\mathbf{d^j \vert v^j})}
\end{align}
As independence is assumed for label decisions in classifying processes, it yields. 
\begin{align}
p(\mathbf{c^j \vert d^j,v^j}) = \frac{1}{f(\mathbf{d^j \vert v^j)}} \prod_{i}{f(\mathbf{d^j} \vert c^j) p(c^j \vert \mathbf{v^j})} 
\end{align}
and the marginalized context-based posterior
\begin{align}
f(c_{i}^{j} \vert \mathbf{d^j,v^j}) \propto f(\mathbf{d_{i}^{j}} \vert c_{i}^{j})p(c_{i}^{j} \vert \mathbf{v_{i}^{j}})
\end{align}
It greatly simplifies MAP posterior estimation since it no longer needs to deal with joint prior conditions of features and contexts. It only needs to obtain two separated likelihoods of the dyadic square given the class value \(C_i\), \(f(\mathbf{d_{i}}^{j} \vert c_{i}^{j})\), and prior context provided through \(\mathbf{v_i}\), \(p(c_{i}^{j} \vert \mathbf{v_{i}^{j}})\). 

While retrieving the likelihood \(f(\mathbf{d_{i}}^{j} \vert c_{i}^{j})\) is straightforward by up-sweeping operations with given HMT model parameters at each scale, the complexity of prior context estimation greatly depends on its structures. Though more general structures may give better prior information for classification, it also greatly complicates the modelling and summarizing information conveyed by \(\mathbf{v_{i}^{j}}\) as well. In other words, we run on the verge of context dilution, especially in case of insufficient training data \cite{cheng1997},\cite{cheng1998},\cite{crouse1998}.

To simplify but still guarantee generalization of prior information, we will employ a simple context structure inspired by the hybrid tree model \cite{bouman1994} in context-labelling trees. Instead of including all neighbouring sub-squares, the simplified context only involves labels from the parent square \(C_{\rho(i)}\) and major vote of class labels from the neighbouring squares \(C_{{\cal N}_{i}}\). As there are only two class labels \(N_c := {0,1}\), the prior context \(\mathbf{v_i} := \{ C_{\rho(i)} , C_{{\cal N}_{i}} \}\) can only been drawn from \(N_{c}^2 = 4\) permuted sets of binary values \(\{0,0\},\{0,1\},\{1,0\},\{1,1\}\). Despite such ad-hoc simple contextual model, it provides sufficient statistic for demonstrating effectiveness of a multi-scale decision fusion process\cite{bouman1994}. Another advantage of the context structure simplification is not requiring enormous number of training data for probability estimation. 

Any decision about labels at a scale \(j\) depends on prior information of labels on a scale \(j-1\); therefore, we can maximize MAP, the equation \ref{eq:map}, in multi-scale coarse-to-fine manner by fusing the likelihood \(f(\mathbf{d_i} \vert c_i)\) given the label tree prior \(p(c_{i}^{j} \vert \mathbf{v_i})\). The fusion step helps pass down MAP estimation through scales to enhance coherency between classifying results of consecutive scales. Moreover, a posterior probability of a class label \(c_i\) given features and the prior context is computed and maximized coherently across multiple scales.
\begin{align}
	\hat{c_i}^{MAP} = argmax_{c_{i}^{j} \in {0,1}} f(c_{i}^{j} \vert \mathbf{d^j,v^j})
\label{eq:map}
\end{align}
\subsection{Multiscale Discriminant Saliency}
\label{subsec:mdis}

Core ideas of DIS and MDIS are measuring discriminant power between two classes centre and surrounds. Though it can be estimated by sample means of mutual information, the underlying mechanism is distinguishing the centre and surround classes given Generalized Gaussian Distribution (GGD) of wavelet coefficients. These distributions are usually zero-mean and well-characterized with only variance parameters. Dashan Gao \cite{gao2009} has estimated scale parameter (variance) of GGD (see section 2.4 \cite{gao2009} for more details) by the maximum a probability process.
\begin{align}
\hat{\alpha}^{MAP} = \left[ \frac{1}{{\cal K}} \left( \sum_{j=1}^{n} \vert x(j) \vert^\beta + \nu \right) \right]^{\frac{1}{\beta}} 
\end{align}
The above MAP formula is then used for deciding whether a sample point or an image data point belongs to either the centre or surround class (see \cite{gao2009} for a detailed proof and explanation). Then, the more distinguishing MAP estimation of the centre class's variance parameter \(\alpha_1\) is from that of the surround class's variance parameter \(\alpha_0\), the more discriminant power for classifying interest from null hypothesis is. 

Beside GGD, GMM is a popular choice for modelling wavelet distributions with variances of multiple classes as well \cite{pesquet1996},\cite{abramovich1998},\cite{moulin1999},\cite{chipman1997}. In binary classification with only two classes, GMM includes two Gaussian Distribution mixtures (GD) of different variances, which are named as "large" / "small" states according to their comparison in terms of variance values. Now the only difference between GMM models and Gao's proposal \cite{gao2007-a} are whether GD or GGD should be used. Though GGD is more sophisticated with customizable distribution shape parameter \(\beta\), several factors support validity of simple GD modelling given the class conditions as hidden variables. Empirical results from estimation have shown that the mixture model is simple yet effective \cite{pesquet1996},\cite{bouman1994}. Modelling wavelet coefficients with hidden classes of "large"/"small" variance states are basic data models in Wavelet-based Hidden Markov Model (HMT) \cite{crouse1998}. With wavelet HMT, image data are processed in a coarse-to-fine multi-scale manner; therefore, MAP of a state \(C_{i}^{j}\) given input features from a sub-square \(D_{i}^{j}\) can be inherently estimated across scales \(j = {0,1,\ldots,J}\). More details about this multi-scale MAP estimation by wavelet HMT can be found in the previous section \ref{subsec:map}. Then, a combination of MAP estimation, in the equation \ref{eq:map} and mutual information computation, in the equation \ref{eq:dis}, yields the MDIS mathematical formulation.
\begin{align}
I_{i}^{j} (C^{j};\mathbf{D^{j}}) = H(C^j)+\sum_{c=0}^{1} P_{C^j \vert \mathbf{D^j}}(c_{i}^{j} \vert \mathbf{d^j}) log P_{C^j \vert \mathbf{D^j}}(c_{i}^{j} \vert \mathbf{d^j})
\label{eq:mdis}
\end{align}
where \(H(C^j) = - p(C^j) log(p(C^j))\) is an entropy estimation of classes across scales \(j\), and the posterior probability can be estimated by modelling wavelet coefficients in the HMT framework. This matter has been discussed in previous sections; therefore, it is not repeated here. As the equation \ref{eq:mdis} yields discriminant power across multiple scales; a strategy is needed for combining them across scales. In this paper, a simple maximum rule is applied for selecting discriminant values from multiple scales into a singular MDIS saliency map at each sub-square \(d_i\). 
\begin{align}
I_{i} (C|\mathbf{D}) = max \left( I_{i}^{j} (C^{j};\mathbf{D^{j}}) \right)
\label{eq:}
\end{align}
\section{Experiments \& Discussion}
\label{sec:exp}
To evaluate saliency representation, reliable ground truth data are utmost necessary. As our research purpose is deepening knowledge about multi-scale discriminant saliency approaches and human visual attention relation, the ground truth data must be gotten from psychological experiments in which human subjects look at different natural scenes and their responses are collectively acquired. Moreover, the research scope only focuses on bottom-up visual saliency, the early stage of attention without interference of top-down knowledge and experiences. Human participants should be naive about aims of experiments and should not know contents of displaying scenes in advance. After these prerequisites are satisfied, human responses on each scene can be accurately collected through eye-tracking equipments. It records collection of eye-fixations for each scene, and these raw data are basic form of ground truths for evaluating efficiency of saliency methods.

In order to standardize the evaluating process, we only utilize one of the most common and accessible database and evaluation tools in visual attention fields in information-based saliency studies, Niel Bruce database \cite{bruce2006}. While proposing his InfoMax (AIM) approach, the first information-based visual saliency, he has simultaneously released its testing database as well. The reasonably small collection with 120 different colour images which are tested by 20 individuals. Each subject observes displayed images in random orders on a 12 inch CRT monitor positioned 0.75 m from their location for 4 seconds with a mask between each pair of images. Importantly, no particular instructions are given except observing the image. 

Above brief description clarified validity of this database for our experiments. DIS \cite{gao2007-a} is obviously the most similar approach to our proposed MDIS, it should as well be used as a reference method. Though the original implementation DIS \cite{gao2007-a} is not available for setting up the comparison, pseudo-DIS can be simulated by MDIS approach with 1x1 window size and zero-mean distribution of pixel values. Besides DIS, the AIM method is also involved as a reference approach against which we compare our proposed saliency solution MDIS in terms of performance, computational load, etc. AIM derives saliency value from information theory with slightly different computation, self-information instead of mutual-information in MDIS or DIS. Therefore, it would be considered the second best as referenced method for our later evaluation of MDIS.

 Besides an appropriate database and referenced methods, proper numerical tools are also necessary for analysing simulation data. As regards to fairness and accuracy, we employ a set of three measurements LCC, NSS, and AUC recommended by Ali Borji et al. \cite{borji2012-c} since evaluation codes can be retrieved freely from their website \footnote{https://sites.google.com/site/saliencyevaluation/}. Rationales behind these evaluation scores ensure reliability of the quantitative observations and conclusions.

In quantitative assessment, general ideas can be drawn about how the proposed algorithms perform in average. However, such evaluation method lacks of specific details about successful and failure cases since the information has been averaged in quantitative method. In an effort of looking for pros and cons of the algorithm, we perform qualitative evaluations for saliency maps generated by MDIS in multiple scales. Furthermore, AIM,DIS saliency maps are generated and compared with MDIS maps for advantages and disadvantages of each method.
\subsection{Quantitative Evaluation}
After the above review of how simulations are built and evaluated in the previous section, following are data representation and analysis of the conducted experiments. In this paper, six dyadic scales are deployed for any HMT training and evaluation; therefore, we have simulation modes from (U/T/V)HMT(0-6) of MDIS depending on whether training stages are deployed (T/V)HMT or universal parameters are used (UHMT). Besides U/T/V prefixes, we also have suffixes (1-5) for consecutive deployments of block sizes from 32x32 down to 2x2. Saliency maps could be combined according to the maximization of mutual information rule, the equation \ref{eq:mdis}; therefore, we have an additional suffix 0 for saliency maps which are created by the across-scale integration. Noteworthy that, whenever a 1x1 window size (a pixel) is used, generated saliency maps are also considered as pseudo DIS saliency maps since MDIS with a pixel-size window is principally equivalent to DIS method in term of resolutions\cite{gao2007}. We assign (U/T)HMT6 to pseudo-DIS saliency maps, as replacements for the original DIS. In addition to DIS, AIM is involved in simulations as the reference method and LCC, NSS, AUC and TIME are chosen as numerical evaluation tools. Below are three tables of simulation results. Table \ref{tab:uhmt} shows experimental data of all UHMT modes while table \ref{tab:thmt} summarizes data of all THMT modes, and table \ref{tab:vhmt} list evaluation results of all VHMT modes.
\begin{table}[!htbp]
\caption{UHMT - MDIS - DATA}
\begin{center}
\resizebox{0.9\textwidth}{!}{
\begin{tabular}{|l|c|c|c|c|}
\hline
	Observations & LCC & NSS & AUC & TIME \\ \hline
	UHMT0 & 0.01434 & 0.21811 & \textbf{0.89392} & 0.39617 \\ \hline
	UHMT1 & \texttt{-0.00269} & 0.19772 & \texttt{0.53862} & 0.39617 \\ \hline
	UHMT2 & 0.01294 & 0.27819 & 0.60520 & 0.39617 \\ \hline
	UHMT3 & 0.01349 & 0.32868 & 0.69065 & 0.39617 \\ \hline
	UHMT4 & \textbf{0.01604} & \textbf{0.42419} & 0.83615 & 0.39617 \\ \hline
	UHMT5 & 0.00548 & \texttt{0.13273} & \textbf{0.89234} & 0.39706 \\ \hline
	UHMT6 (DIS) & 0.00553 & 0.46003 & 0.85658 & 0.88706 \\ \hline
\end{tabular}
}
\end{center}
\label{tab:uhmt}

\caption{THMT - MDIS - DATA}
\begin{center}
\resizebox{0.9\textwidth}{!}{
\begin{tabular}{|l|c|c|c|c|}
\hline
	Observations & LCC & NSS & AUC & TIME \\ \hline
	THMT0 & \textbf{0.02382} & \textbf{0.48019} & \textbf{0.88357} & 2.32734 \\ \hline
	THMT1 & \textbf{0.02582} & 0.38096 & \texttt{0.60922} & 2.32734 \\ \hline
	THMT2 & 0.01156 & 0.31855 & 0.64633 & 2.32726 \\ \hline
	THMT3 & 0.01604 & 0.32491 & 0.71972 & 2.32726 \\ \hline
	THMT4 & 0.01143 & \texttt{0.29662} & 0.81192 & 2.32726 \\ \hline
	THMT5 & \texttt{0.00512} & 0.36932 & \textbf{0.89532} & 2.32726 \\ \hline
	THMT6 (DIS) & -0.2673 & 0.13989 & 0.83324 & 7.28872 \\ \hline
\end{tabular}
}
\end{center}
\label{tab:thmt}

\caption{VHMT - MDIS - DATA}
\begin{center}
\resizebox{0.9\textwidth}{!}{
\begin{tabular}{|l|c|c|c|c|}
\hline
Observations & LCC & NSS & AUC & TIME \\ \hline
VHMT0 & 0.01697 & \textbf{0.44170} & \textbf{0.86606} & 2.84212 \\ \hline
VHMT1 & 0.01693 & 0.38387 & \texttt{0.61187} & 2.84212 \\ \hline
VHMT2 & \textbf{0.02044} & 0.38777 & 0.67060 & 2.84212 \\ \hline
VHMT3 & 0.01430 & 0.38882 & 0.73682 & 2.84212 \\ \hline
VHMT4 & 0.00946 & \texttt{0.36761} & 0.82329 & 2.84212 \\ \hline
VHMT5 & \texttt{-0.00125} & 0.39580 & \textbf{0.88160} & 2.84212 \\ \hline
AIM & 0.01576 & 0.12378 & 0.72400 & 50.41714 \\ \hline
\end{tabular}
}
\end{center}
\label{tab:vhmt}
\end{table}
\ifissvjour
\begin{figure*}[!htbp]%
\fi
\ifiselsarticle
\begin{figure}[!htbp]%
\fi
\ifisreport
\begin{figure}[!htbp]%
\fi
\centering
\begin{subfigure}[b]{0.48\textwidth}
	\includegraphics[width=\textwidth,height=0.3\textheight]{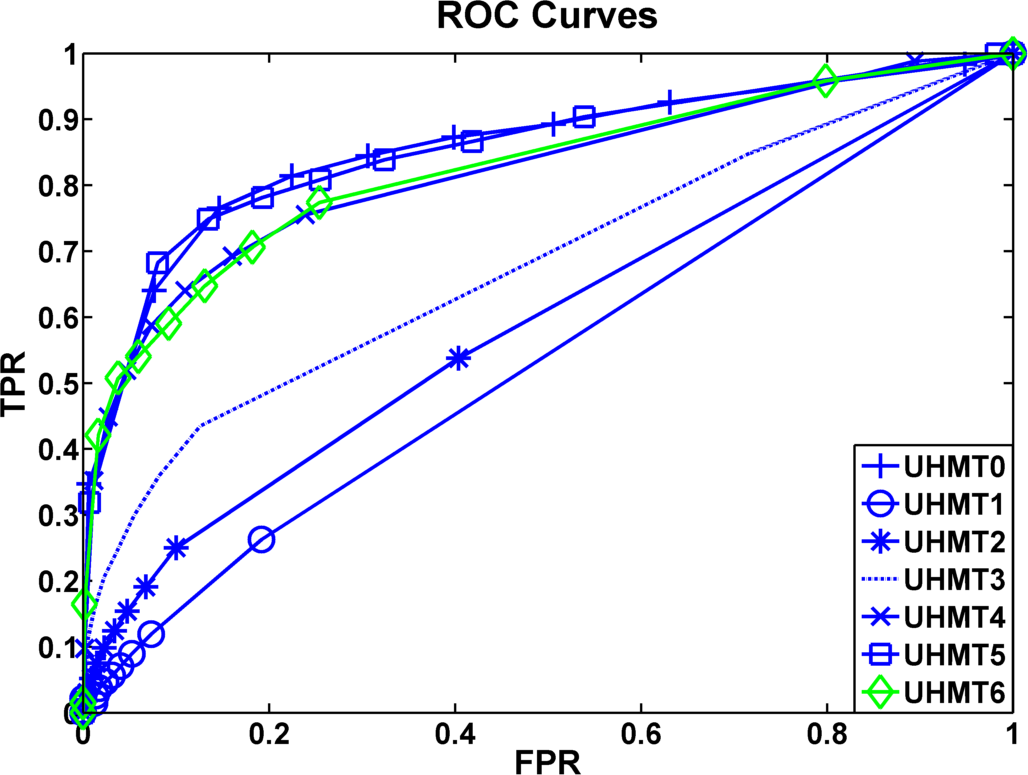}	
	\caption{UHMT - MDIS - ROC}
	\label{fig:uhmt_roc}
\end{subfigure}
\quad
\begin{subfigure}[b]{0.48\textwidth}
	\includegraphics[width=\textwidth,height=0.3\textheight]{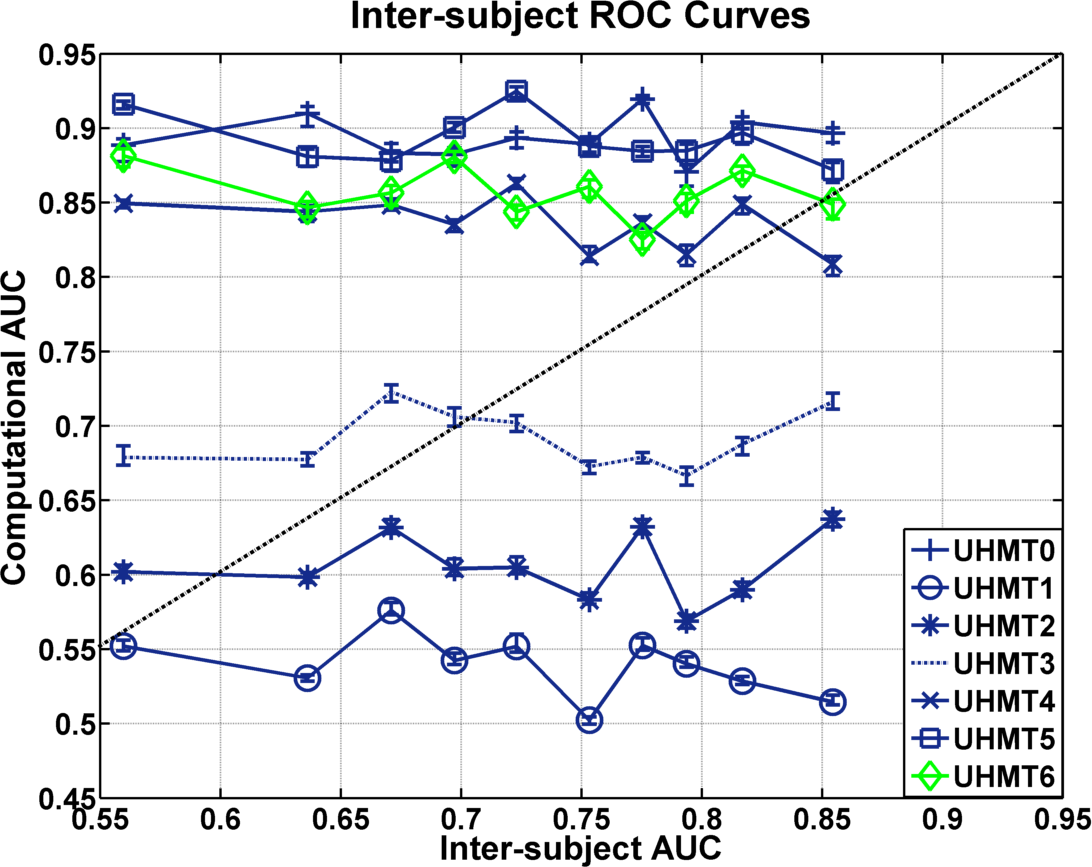}	
	\caption{UHMT - MDIS - ISROC}
	\label{fig:uhmt_isroc}
\end{subfigure}
\\
\begin{subfigure}[b]{0.48\textwidth}
	\includegraphics[width=\textwidth,height=0.3\textheight]{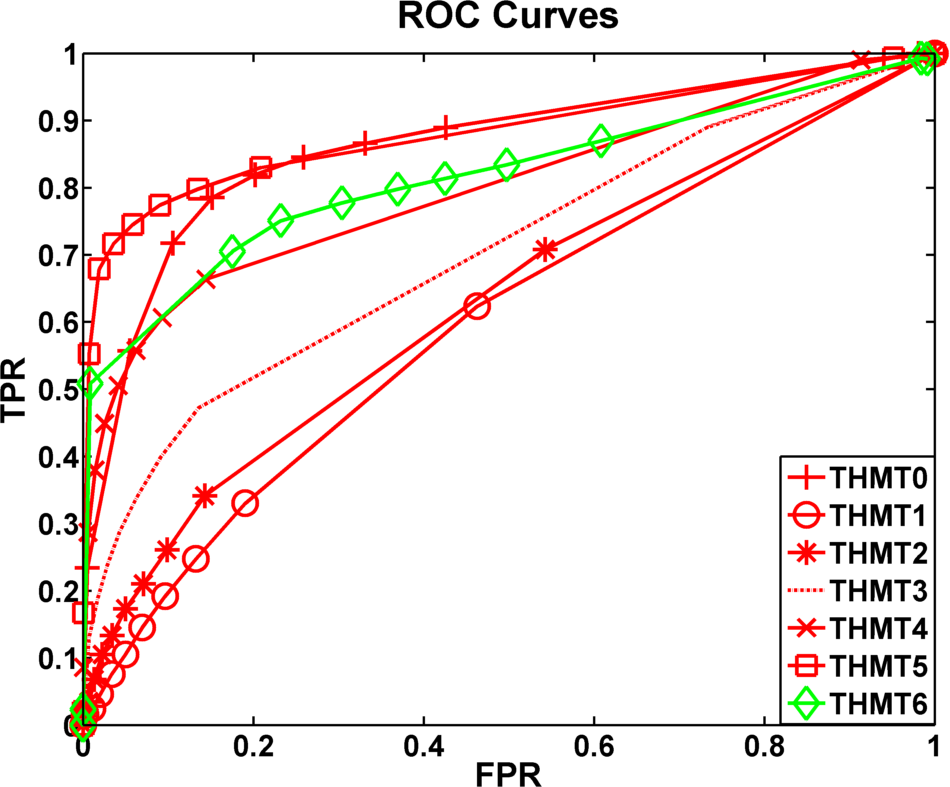}	
	\caption{THMT - MDIS - ROC}
	\label{fig:thmt_roc}
\end{subfigure}
\quad
\begin{subfigure}[b]{0.48\textwidth}
	\includegraphics[width=\textwidth,height=0.3\textheight]{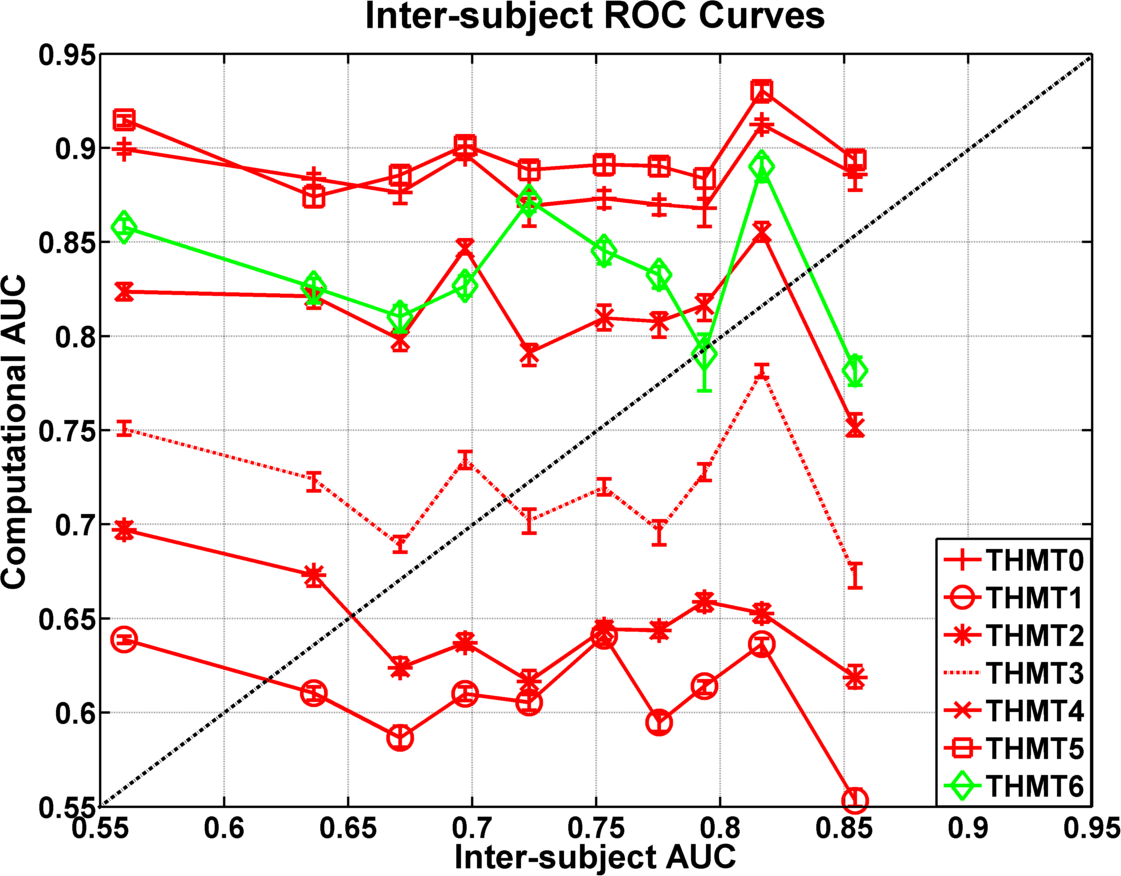}	
	\caption{THMT - MDIS - ISROC}
	\label{fig:thmt_isroc}
\end{subfigure}
\\
\begin{subfigure}[b]{0.48\textwidth}
	\includegraphics[width=\textwidth,height=0.3\textheight]{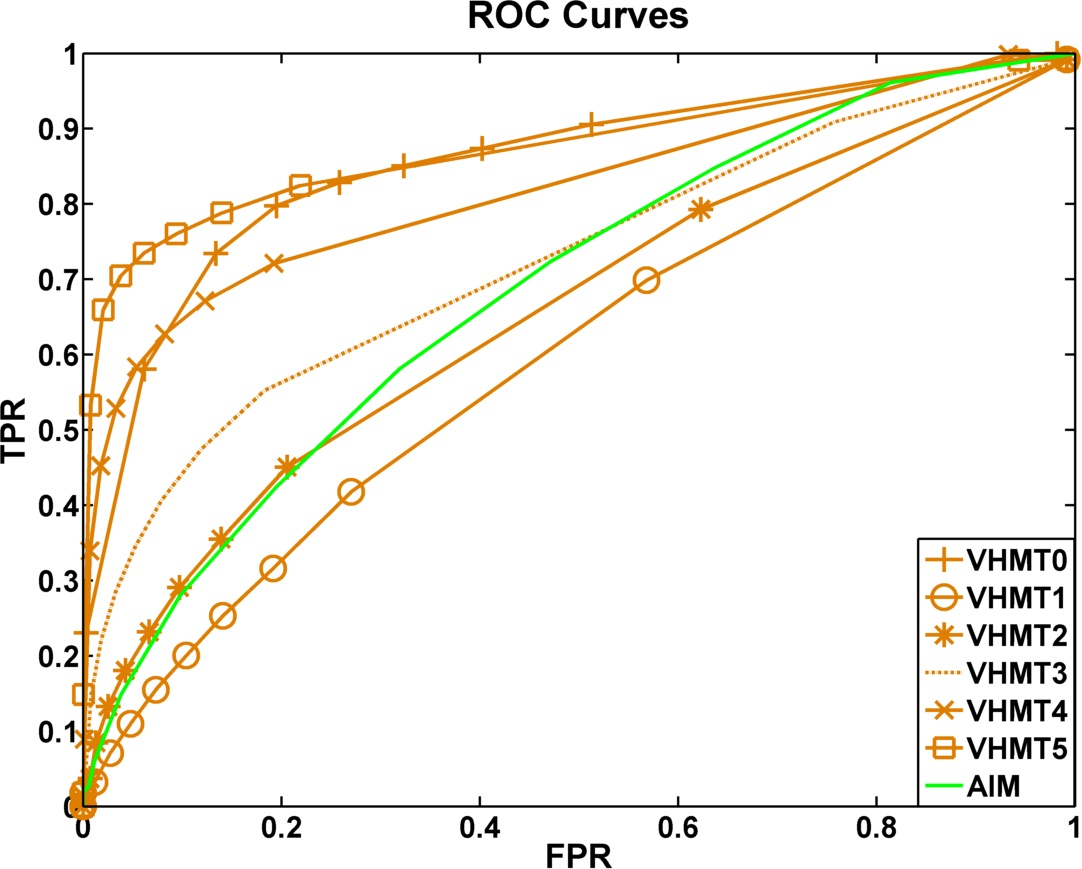}	
	\caption{VHMT - MDIS - ROC}
	\label{fig:vhmt_roc}
\end{subfigure}
\quad
\begin{subfigure}[b]{0.48\textwidth}
	\includegraphics[width=\textwidth,height=0.3\textheight]{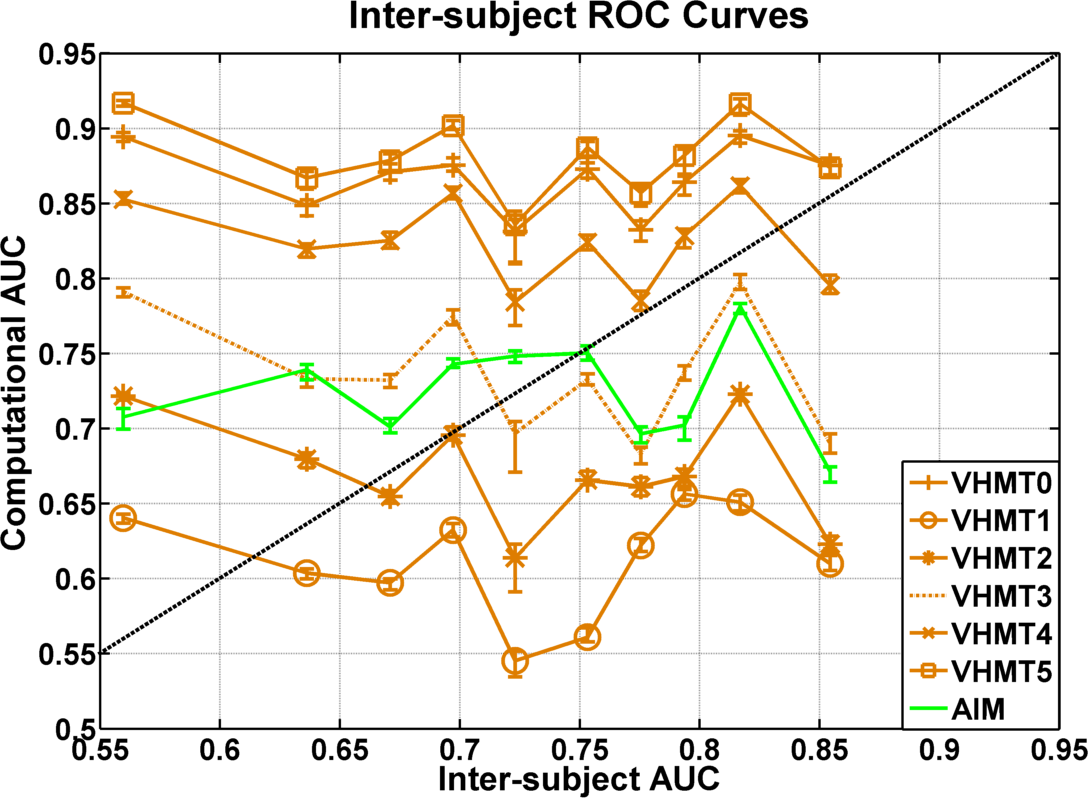}	
	\caption{VHMT - MDIS - ISROC}
	\label{fig:vhmt_isroc}
\end{subfigure}
\caption{(U/T/V)HMT - MDIS - (IS)ROC}
\label{fig:rocs}
\ifisreport
\end{figure}
\fi
\ifiselsarticle
\end{figure}
\fi
\ifissvjour
\end{figure*}
\fi

Among above numerical evaluation tools for visual saliency, Receiver Operating Curve (ROC) and its Area Under Curve (AUC) are the most popular. It measures efficiency of saliency maps in classifying fixation and non-fixation points of human eye movements in visual psychological experiments. In ROC curve, the vertical axis indicates True Positive Rate of the classification which is equal to hit rate, recall measurement. It is the ratio between correctly classified fixation points and its total number. Meanwhile, False Positive Rate (FPR) is equivalent with a fall-out ratio, a number of incorrectly classified fixation points over a total number of non-fixation points. 
Figures \ref{fig:uhmt_roc},\ref{fig:thmt_roc},\ref{fig:vhmt_roc} display ROC curves of UHMT, THMT, VHMT for several scales with AIM as reference curves. In all figures, solid green lines are representing ROCs of referenced methods such as AIM or DIS (HMT6); while, blue, red, and orange colours represent ROCs of UHMT, THMT, and VHMT consequently. In general, AIM and DIS have moderate performances when compared with the proposed multi-scale HMT. AIM and DIS perform better than HMT with large sampling windows. When smaller blocks are employed, HMTs surpass AIM and DIS in detection of fixation points. The order of ROC performance between AIM, DIS and HMTs methods can be summarized as follows. 
\begin{alignat}{7}
HMT5 &\geq HMT0 &> DIS &\geq HMT4 \space &\geq AIM &> HMT3 &> HMT2 &> HMT1
\label{eq:cmpord}
\end{alignat}

In comparisons of (U,T,V) HMTs, we can see advantages of (T,V)HMT over UHMT on different scales. It is clearly shown in three corresponding ROC figures \ref{fig:uhmt_roc},\ref{fig:thmt_roc},\ref{fig:vhmt_roc} since curves of (T,V)HMT moves closer toward the left-top corner than UHMT does. It means (T,V)HMT more successfully detect meaningful points than UHTM does in region of low fall-out rate (FPR). Therefore, (T,V)HMT provides better binary classifier in term of robustness. It is reasonable since THMT, VHMT requires training steps on image data meanwhile UHMT model just uses predefined and general parameters. Furthermore, VHMT has slightly better performance than THMT in most window sizes. The slight increment in terms of ROC curves is due to the fact VHMT is a more comprehensive model than THMT when processing texture features \cite{do2002}.

Generalized ROC evaluates performances of saliency maps with different thresholds, according to ground-truth eye-fixation data of several human test subjects. However, it does not distinguish eye-fixation data from each subject but treats the data collectively. Therefore, it certainly loses important aspects of eye-tracking data from multiple subjects such as diversity of individual responses over various types of scenes shown during experiments. 
To assess limitation of ROC evaluation, we take suggestion of Harel \cite{harel2007} by applying the ISROC evaluation method. The correspondent simulation generates graphs \ref{fig:uhmt_isroc},\ref{fig:thmt_isroc}, and \ref{fig:vhmt_isroc} for (UHMT) universal, (THMT) trained and (VHMT) vector-based configuration. Generally, all saliency methods have quite consistent performances across different complexity of scenes. In other words, they could produce meaningful saliency maps in complex cases where inter-subject scores are low, human subjects are struggling to find common salient points. However, these computational saliency maps are overcomplicate in simple scenes with high inter-subject ROC scores. In these cases, all subjects focus on a few locations on testing scenes, while the proposed saliency methods still detect other regions of interest beside main objects. Therefore, the proposed computational approaches are less efficient than human beings in such situations.

Similar to ROC, ISROC curves help to compare HMTs with different configurations on multiple levels as well. Observing figure \ref{fig:rocs} shows advantages of HMTs in terms of ISROC curves when smaller block sizes are chosen. Moreover, the reference method AIM's performance, the solid green line, is better than HMT1 (32x32 blocks) and HMT2 (16x16 blocks), equivalent with HMT3 (8x8 blocks), but worse than HMT0, HMT4 (4x4 blocks), and HMT5 (2x2 blocks). Meanwhile, DIS is better than HMT1-3, almost equivalent to HMT4 and worse than HMT0 and HMT5. The ranking orders of HMT\#, AIM and DIS in terms of ISROC are almost similar the rank in terms of ROC, the mathematical comparison \ref{eq:cmpord}. HMT0, HMT4 and HMT5 have better performance than human subjects in most of situation; their ISROC curves are above the black boundary set by  human subjects for most of the plots. Noteworthy, there is an interesting observation about HMT0 - the integrated map, and HMT5 - 2x2 blocks that HMT5 is almost equivalent with or sometimes better than the integrated method HMT0. Perhaps, more complex rules for integrating saliency maps need developing to efficiently integrate attention maps with different scales.

Plots of ROC and ISROC curves give general ideas that HMTs overperform the referenced AIM and DIS maps; however, it does not specify how much better the proposed methods are. Hence, it is necessary to have other numerical analysis of their performances. Computational loads are the first measure to be analysed and compared among saliency approaches. In the TIME rows of tables \ref{tab:uhmt},\ref{tab:thmt} and \ref{tab:vhmt}, we present necessary processing time for each method or each mode. Generally, computational loads, proportional to processing time, of all modes in either UHMT or THMT row is almost similar since the parameters of full-depth Hidden Markov Tree need estimating before computation of saliency values. In comparison of (U/T/V) HMTs in terms of processing time, UHMT is faster than (T/V)HMT as UHMT uses predefined parameters instead of learning HMT parameters from each image. When comparing (U/T/V) HMT modes of MDIS with AIM, our proposed methods are much faster than AIM. The well-known AIM directly estimates self information from high-dimensional by ICA algorithm while MDIS statistically models two hidden states: "large" \ "small" states in sparse and structural features with efficient and fast inference algorithms. Computational load or processing time of the mentioned AIM and proposed MDIS with different modes can be seen in the figure \ref{fig:time}.
\ifissvjour
\begin{figure*}[!htbp]
\fi
\ifiselsarticle
\begin{figure}[!htbp]
\fi
\ifisreport
\begin{figure}[!htb]
\fi
	\centering
	\begin{subfigure}[b]{0.425\textwidth}
		\includegraphics[width=\textwidth]{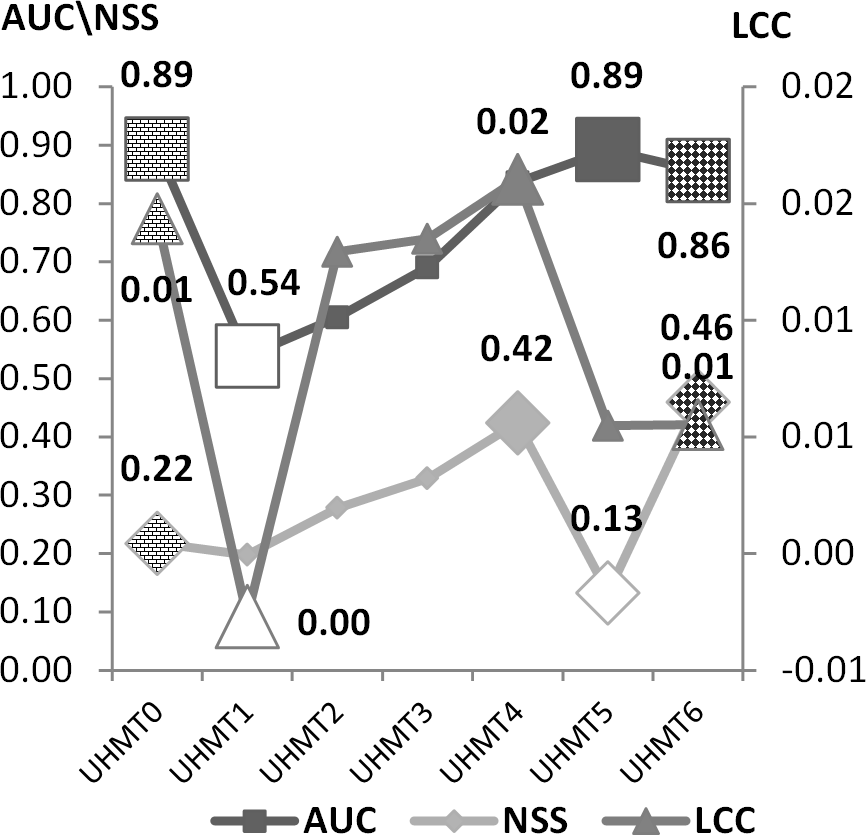}
		\caption{UHMT - MDIS}
		\label{fig:uhmt}
	\end{subfigure}	
	\quad
	\begin{subfigure}[b]{0.425\textwidth}
		\includegraphics[width=\textwidth]{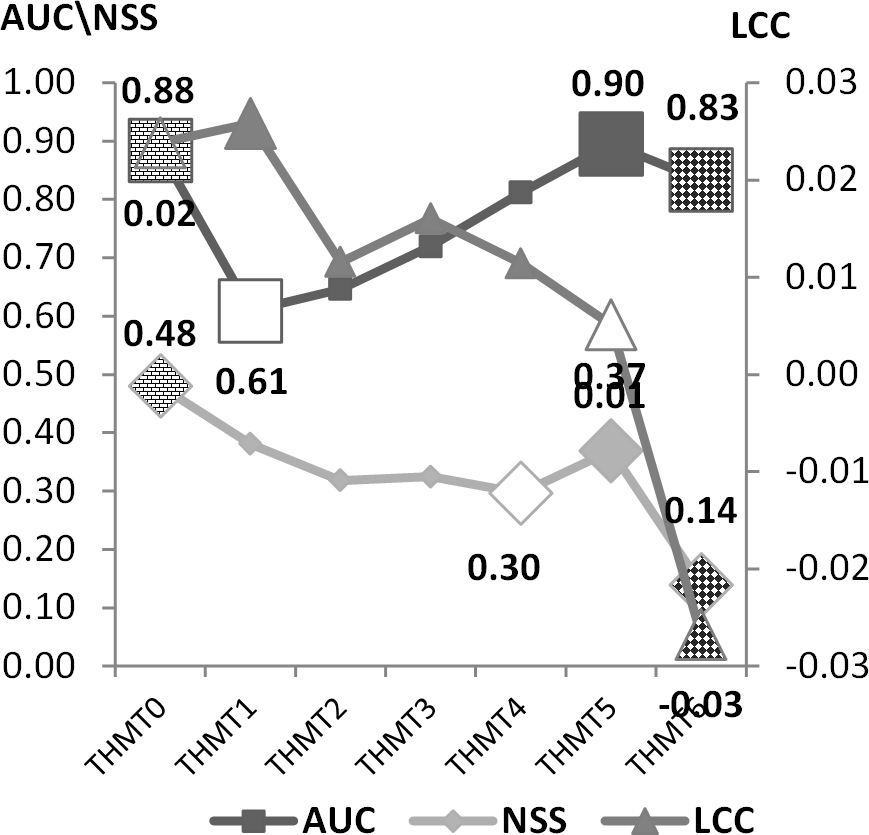}
		\caption{THMT - MDIS}
		\label{fig:thmt}
	\end{subfigure}	
	\\	
	\begin{subfigure}[b]{0.425\textwidth}
		\includegraphics[width=\textwidth]{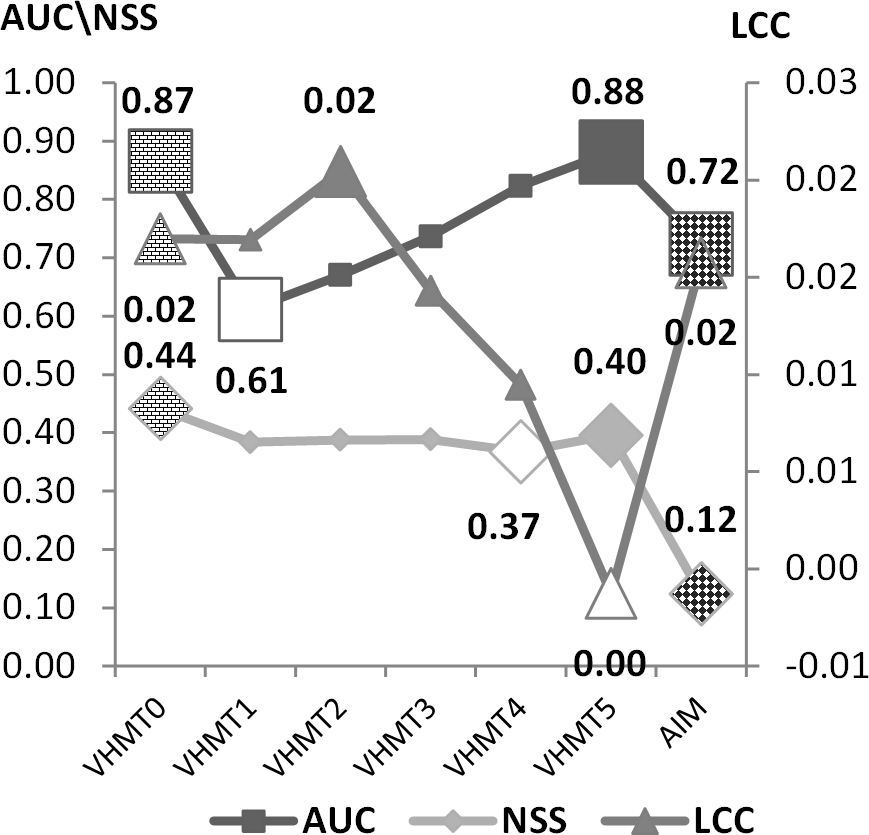}
		\caption{VHMT - MDIS}
		\label{fig:vhmt}
	\end{subfigure}	
	\quad
	\begin{subfigure}[b]{0.425\textwidth}
		\includegraphics[width=\textwidth]{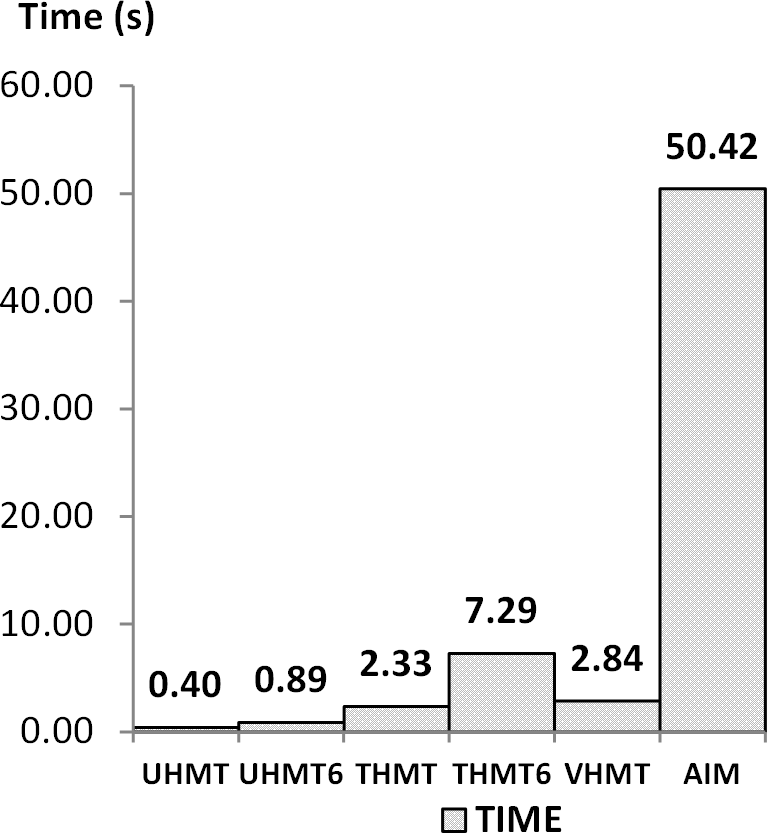}
		\caption{TIME}
		\label{fig:time}
	\end{subfigure}		
	\caption{Performance of UHMT-MDIS, THMT-MDIS in AUC,NSS,LCC and TIME}
	\label{fig:hmtexpres1}
\ifisreport
\end{figure}
\fi
\ifiselsarticle
\end{figure}
\fi
\ifissvjour
\end{figure*} 
\fi
Though HMT-MDIS significantly reduces computational load for computation of information-based saliency, more verifications are necessary for their performances in terms of accuracy. We begin evaluating three modes (U/T/V)HMT separately against AIM and pseudo-DIS in terms of three numerical tools LCC, NSS, AUC together, Figure \ref{fig:uhmt}, \ref{fig:thmt}, \ref{fig:vhmt}. Then all modes of HMTs are summarized in three plots ( the top row of Figure \ref{fig:summary} ) in the following order NSS, LCC, AUC from left to right. Especially in the figure \ref{fig:summary}, simulation modes of the same scale level are placed next to each other for example (U/T/V) HMT0 sit next to each other, so do (U/T/V) HMT1 and etc. It is intentionally arranged in that way to compare performances of different simulation modes in the same scale level. Noteworthy that, in both tables \ref{tab:uhmt} and \ref{tab:thmt} for each row is identified \textbf{maximum} and \texttt{minimum} values by corresponding text styles. Identification of extreme values only involves derivatives (U,T,V) HMT of MDIS modes. In the figures \ref{fig:uhmt}, \ref{fig:thmt}, \ref{fig:vhmt}, extreme (maximum or minimum) values are also specially marked. For example, maximum values have big solid markers while big ones without any textures represent minimum points and marks with "brick-wall" textures are for integrated HMT methods. Especially, AIM and HMT6 (DIS) have big markers with distinguishing big cross-board texture while integrated saliency modes of (U/T/V)HMT0 have small cross-board textures. These special markers help highlight interesting comparisons of the proposed MDIS against AIM and DIS. The same marking policy is applied for data representation in the figure \ref{fig:summary}. Meanwhile, each line in this figure has an arrow head for showing trends of experimental data (increasing/decreasing) when simulation modes are changed across U,T, or V configurations of HMT for each scale level.

\begin{sidewaysfigure}[!htbp]
\includegraphics[width=\textwidth,height=0.7\textheight]{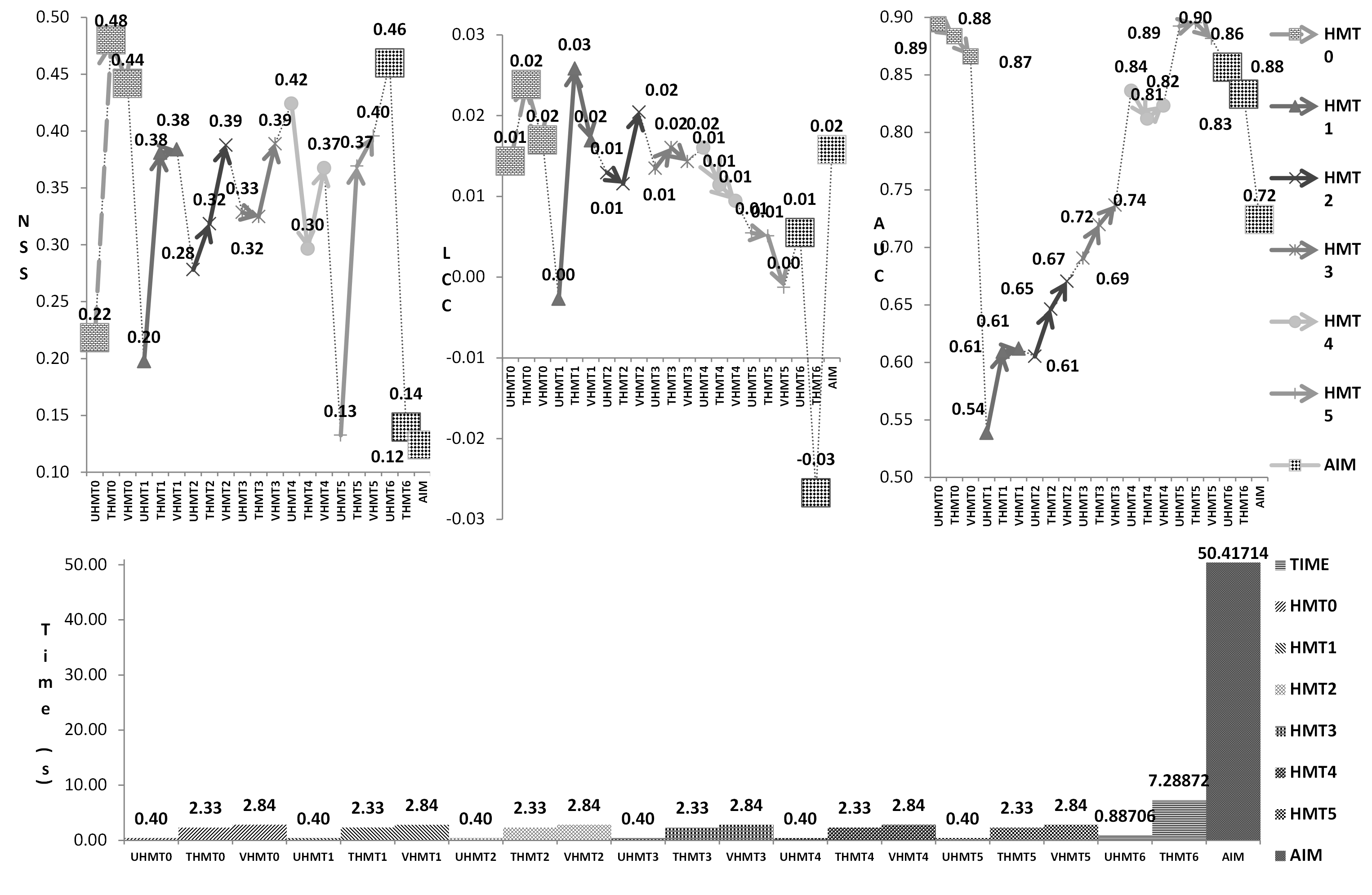}
\caption{Summary of all MDIS against AIM/DIS}%
\label{fig:summary}%
\end{sidewaysfigure}

According to figures \ref{fig:time} and \ref{fig:summary}, UHMT modes require very little effort in saliency computation. Obviously, that fact raises a question about its accuracy of centre and surround classifier as well as synthesized saliency maps. According to simulation data in the table \ref{tab:uhmt} with highlighted extrema, UHMT performs pretty well against AIM/DIS in all three measurements LCC, NSS and AUC. For example, MDIS with UHMT4 mode ( 4x4 square blocks ) surpasses AIM in all measurements. It confirms validity and efficiency of our proposed methods in the information-based saliency map research field. When performances of different UHMT-MDIS modes are considered, UHMT4 with 4x4 squares  have the most consistent evaluation among all dyadic scales with maximum LCC and NSS and the second best AUC value. UHMT0-MDIS, integration of saliency values across scales, does not have better performance than other UHMTs except for AUC level. It shows inconsistent side of deploying HMT with predefined universal parameters which requires no training effort for adapting the model into multi-scale statistical structures.

Different from UHMT mode, training stages are included in the simulation of MDIS with (T/V) HMT mode. With additional adaptivity, (T/V)HMT might improve the saliency evaluation and produce more consistent results than UHMT might. This subjection is solidified by simulation data in the tables \ref{tab:thmt},\ref{tab:vhmt} and they are also plotted in the figures \ref{fig:thmt},\ref{fig:vhmt}. As observed in these tables, all \textbf{maximum} values locate at the THMT0 column, THMT0-MDIS over-performs AIM/DIS in all evaluating schemes. Again, the rationale of MDIS is confirmed and proved by experimental results. Furthermore, effectiveness of training stages is clearly shown when comparing THMT0 against UHMT0. Though AUC of THMT0 is smaller than that of UHMT0, other evaluations of THMT0 are better than their counterparts in both NSS and LCC schemes. This confirms usefulness of training Hidden Markov Tree models for each sample image. In addition, the figure \ref{fig:thmt},\ref{fig:vhmt} shows supremacy of (T,V) HMT0 modes, the across-scale integration mode of MDIS over other singular saliency maps at different dyadic scales in any measurement. Noteworthy, that LCC of THMT0 mode is a bit smaller than LCC of THMT1 mode; however, this small difference can be safely ignored. Comparison of (U/T/V)HMT-MDIS mode-by-mode between data in the table \ref{fig:uhmt},\ref{fig:thmt},\ref{fig:vhmt} are shown in the figure \ref{fig:summary}. Accordingly, there are slight improvements of (T/V)HMT1,(T/V)HMT2 over UHMT1, UHMT2, equivalence of THMT3, UHMT3, and a reverse trend that UHMT4-5 are comparable or slightly better than THMT4, THMT5. It seems that training processes are more important when big classification windows are used. Meanwhile universal approaches of HMT work pretty well if dyadic squares get smaller. Two possible reasons for this observation are statistical natures of dyadic squares and characteristics of training processes. A bigger square has richer joint-distribution of features; therefore, UHMT with fixed parameters can not marginally approximate that distribution well. However, parameters of (T,V)HMT models can be learn from analysing images; it results in significant improvement of saliency maps quality. While smaller sub-squares are less statistically distinguishing, they are successfully modelled by universal parameters of HMT. In these cases, training processes might become redundant since UHMT would perform as well as THMT would do.
\subsection{Qualitative Evaluation}
In this section, saliency maps are analysed qualitatively or visually. From this analysis, we want to identify (i) on which image contexts (U/T/V) HMT-MDIS work well, (ii) how scale parameters affect formation of saliency maps, and (iii) how MDIS in general is compared with AIM/DIS.
\ifiselsarticle
\begin{figure}[!htbp]%
\centering
\begin{subfigure}[b]{0.49\textwidth}
\begin{subfigure}[b]{\textwidth}
	\includegraphics[width=\textwidth,height=0.15\textheight]{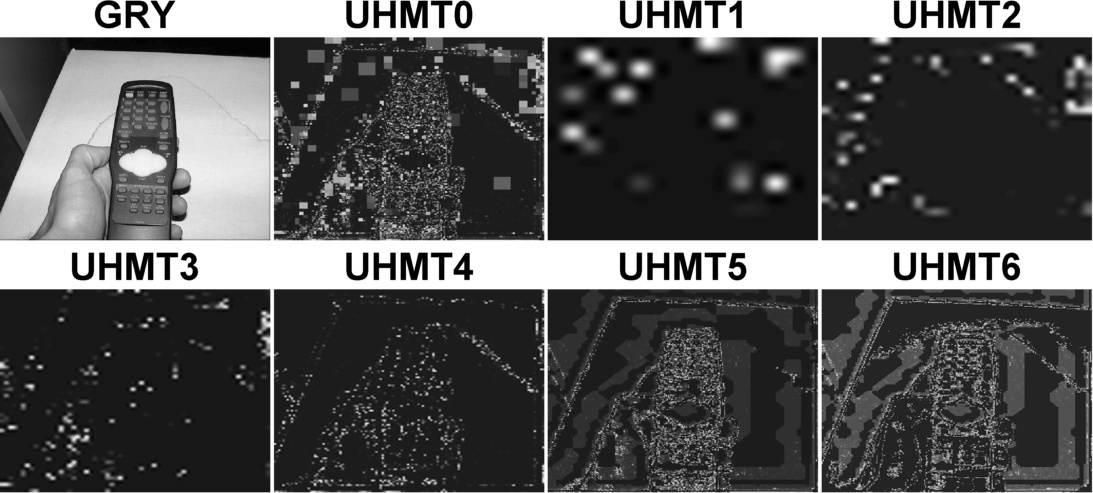}
	\caption{Sample 1 - UHMT}
	\label{fig:uhmt_19}
\end{subfigure}
\\
\begin{subfigure}[b]{\textwidth}
	\includegraphics[width=\textwidth,height=0.15\textheight]{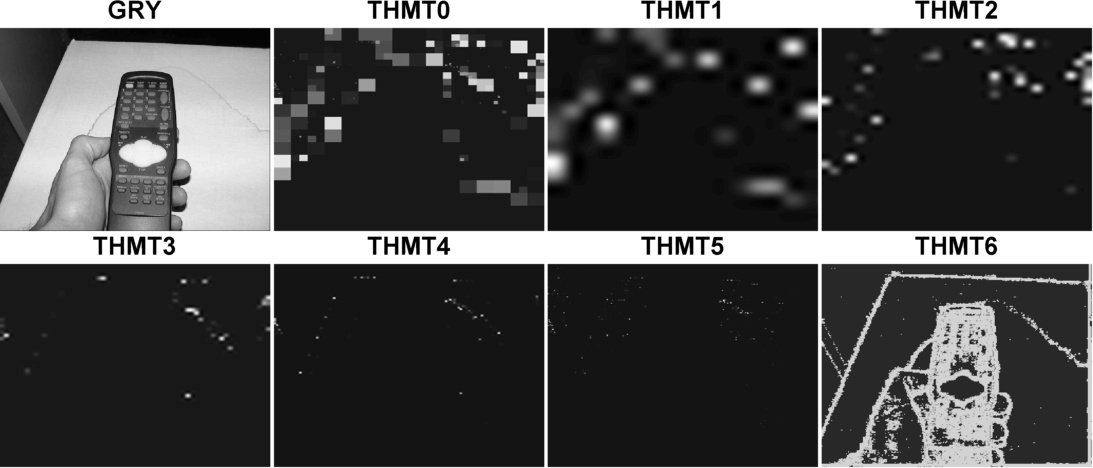}
	\caption{Sample 1 - THMT}
	\label{fig:thmt_19}
\end{subfigure}
\\
\begin{subfigure}[b]{\textwidth}
	\includegraphics[width=\textwidth,height=0.15\textheight]{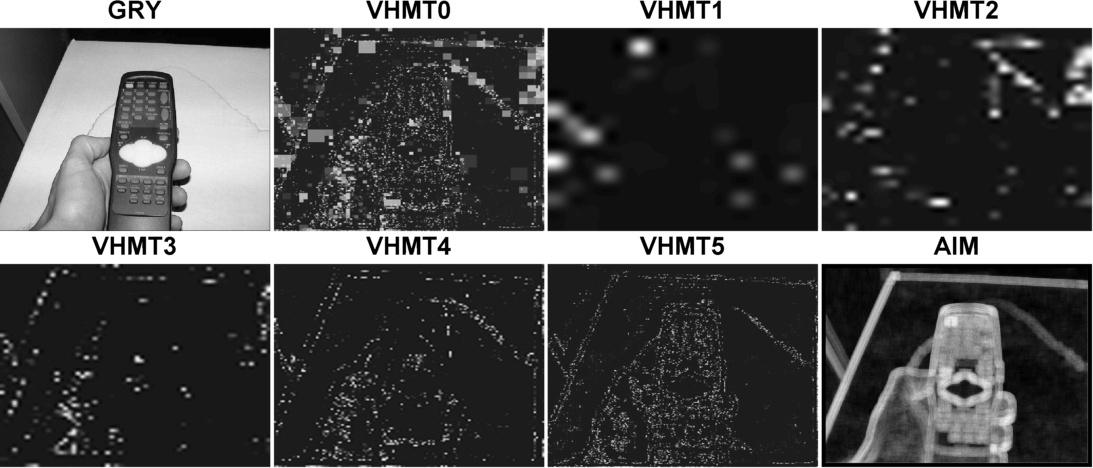}
	\caption{Sample 1 - VHMT}
	\label{fig:vhmt_19}
\end{subfigure}
\end{subfigure}
\begin{subfigure}[b]{0.49\textwidth}%
\begin{subfigure}[b]{\textwidth}
	\includegraphics[width=\textwidth,height=0.15\textheight]{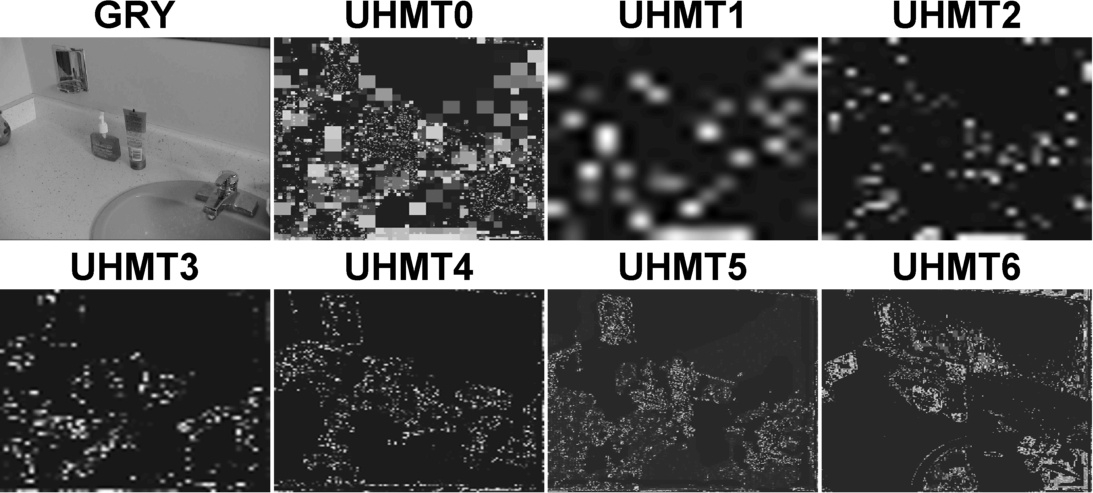}
	\caption{Sample 2 - UHMT}
	\label{fig:uhmt_13}
\end{subfigure}
\\
\begin{subfigure}[b]{\textwidth}
	\includegraphics[width=\textwidth,height=0.15\textheight]{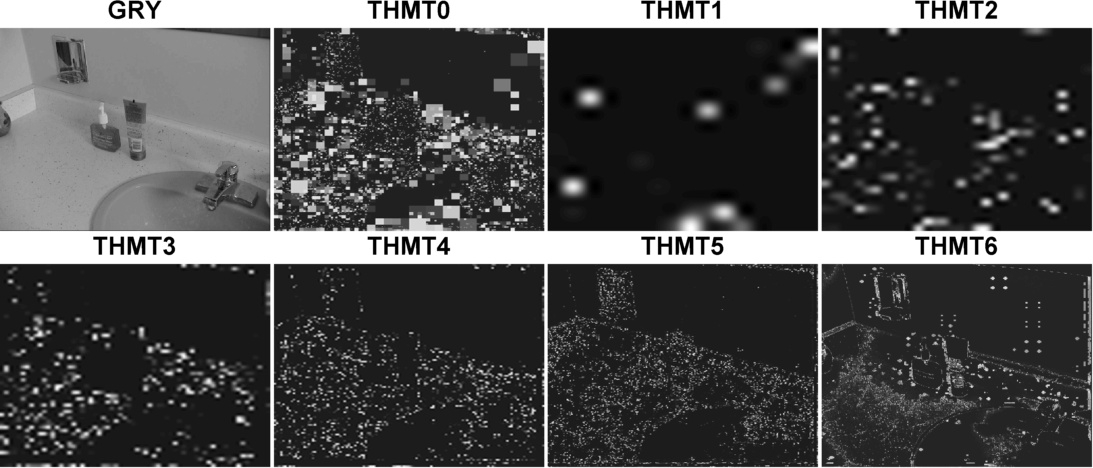}
	\caption{Sample 2 - THMT}
	\label{fig:thmt_13}
\end{subfigure}
\\
\begin{subfigure}[b]{\textwidth}
	\includegraphics[width=\textwidth,height=0.15\textheight]{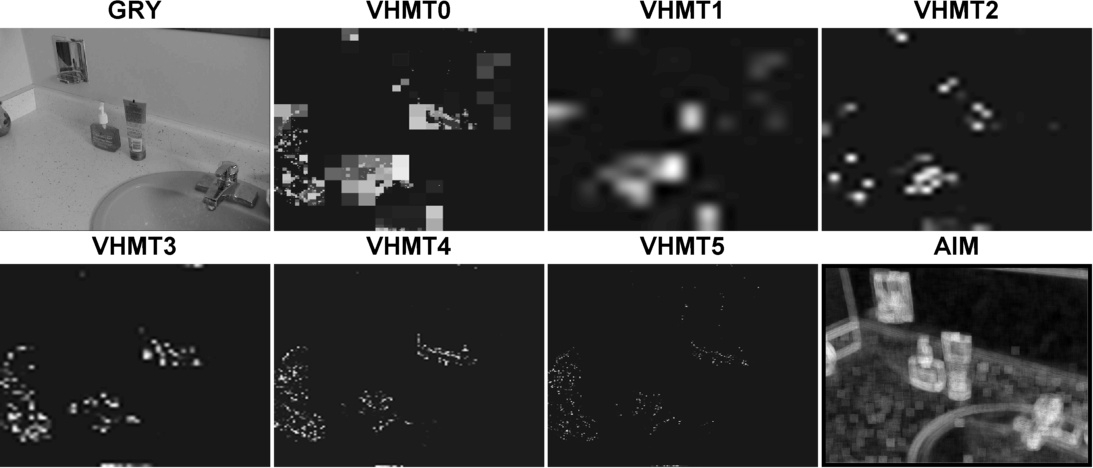}
	\caption{Sample 2 - VHMT}
	\label{fig:vhmt_13}
\end{subfigure}
\end{subfigure}
\\
\begin{subfigure}[b]{0.49\textwidth}
\begin{subfigure}[b]{\textwidth}
	\includegraphics[width=\textwidth,height=0.15\textheight]{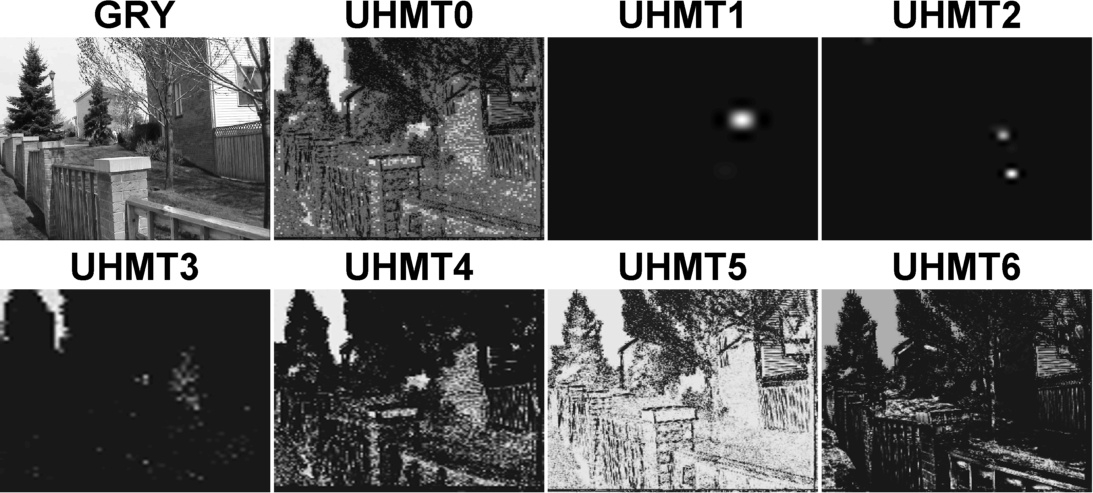}
	\caption{Sample 3 - UHMT}
	\label{fig:uhmt_102}
\end{subfigure}
\begin{subfigure}[b]{\textwidth}
	\includegraphics[width=\textwidth,height=0.15\textheight]{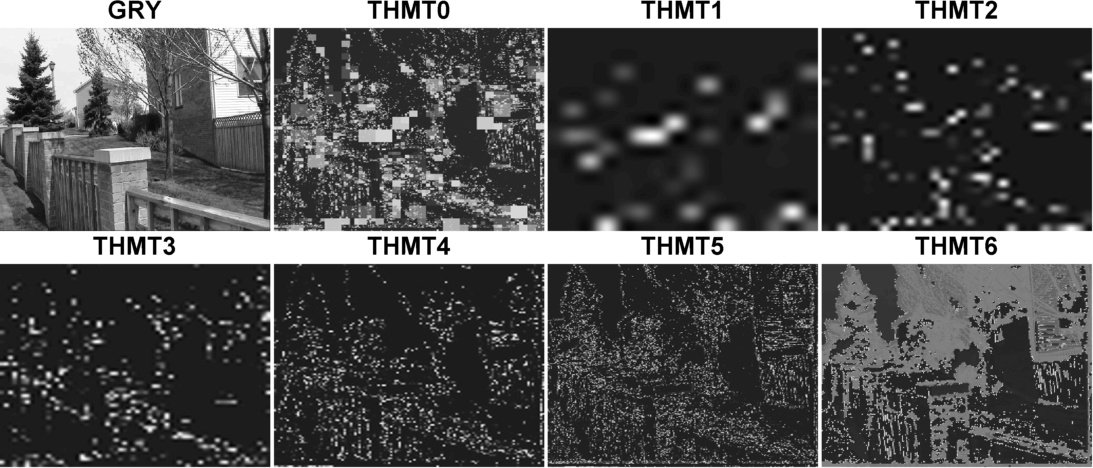}
	\caption{Sample 3 - THMT}
	\label{fig:thmt_102}
\end{subfigure}
\\
\begin{subfigure}[b]{\textwidth}
	\includegraphics[width=\textwidth,height=0.15\textheight]{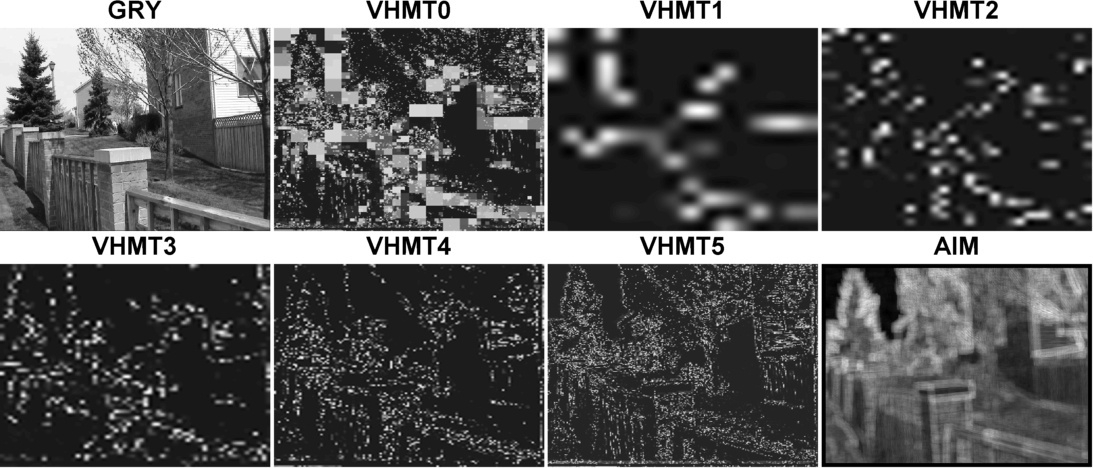}
	\caption{Sample 3 - VHMT}
	\label{fig:vhmt_102}
\end{subfigure}
\end{subfigure}
\begin{subfigure}[b]{0.49\textwidth}
\begin{subfigure}[b]{\textwidth}
	\includegraphics[width=\textwidth,height=0.15\textheight]{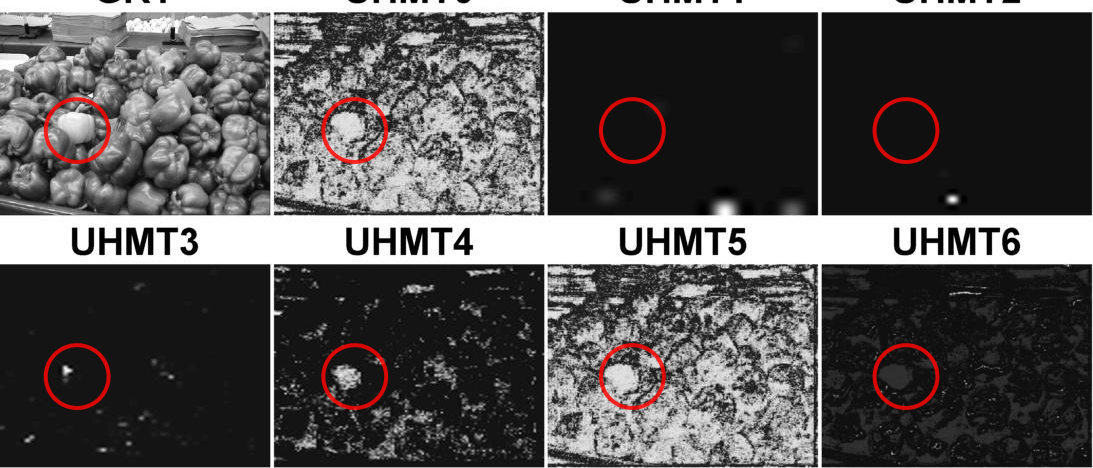}
	\caption{Sample 4 - UHMT}
	\label{fig:uhmt_111}
\end{subfigure}
\\
\begin{subfigure}[b]{\textwidth}
	\includegraphics[width=\textwidth,height=0.15\textheight]{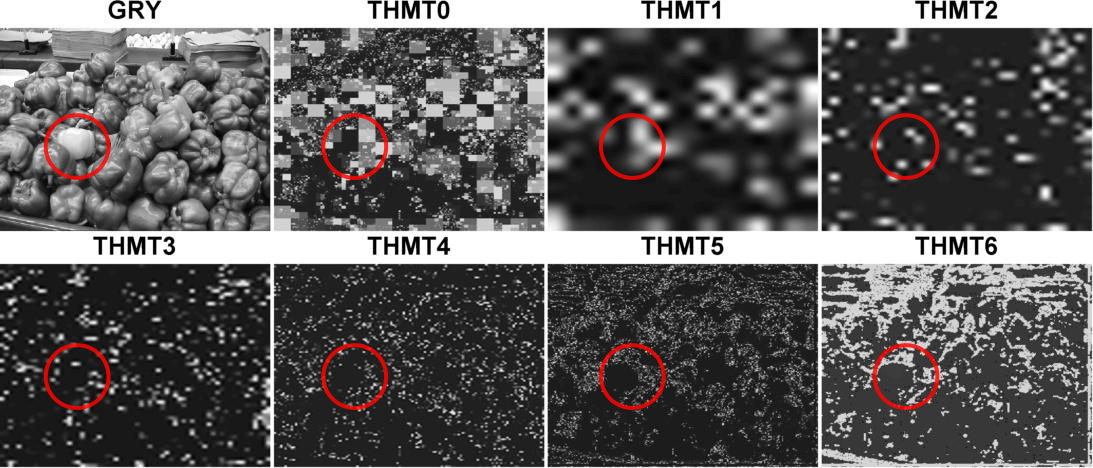}
	\caption{Sample 4 - THMT}
	\label{fig:thmt_111}
\end{subfigure}
\\
\begin{subfigure}[b]{\textwidth}
	\includegraphics[width=\textwidth,height=0.15\textheight]{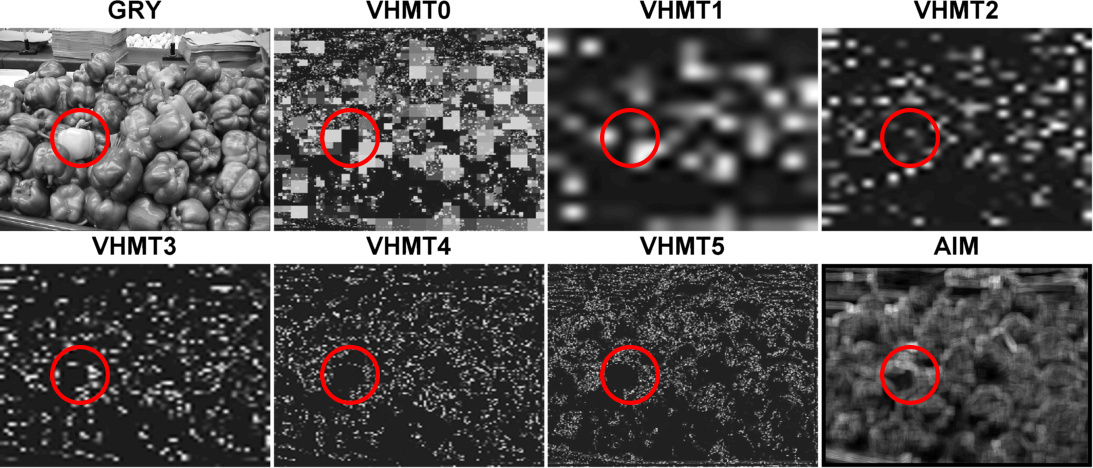}
	\caption{Sample 4 - VHMT}
	\label{fig:vhmt_111}
\end{subfigure}
\end{subfigure}
\caption{Saliency Maps}%
\label{fig:expdis1}%
\end{figure}
\fi
\ifisreport
\else
In figures \ref{fig:uhmt_19},\ref{fig:thmt_19},\ref{fig:vhmt_19}, and \ref{fig:uhmt_13},\ref{fig:thmt_13},\ref{fig:vhmt_13}, sample images with
\fi
 big central objects show an example of good (U/V)HMT performance but bad THMT performance. All scale levels of THMT suppress features of the most obvious objects in the image centre. Meanwhile, (U/V)HMT4 and (U/V)HMT5 capture significant features of those objects; therefore, (U/V)HMT0 has much better saliency map than THMT0. In this case, the best saliency map of MDIS approach, (U/V)HMT0, is reasonably competitive against AIM/DIS. Despite different configurations, UHMT-MDIS and VHMT-MDIS have quite similar results for simple scenes with few central objects. Observing the figures \ref{fig:uhmt_19},\ref{fig:uhmt_13} and \ref{fig:vhmt_19},\ref{fig:vhmt_13}, we can see differences between generated saliency maps of UHMT and VHMT across six different modes [0-5] especially mode 5. This mode utilizes the 2x2 blocks; therefore, it can create saliency maps with great details about discrepant regions in (U/V)HMT modes. Generally, UHMT tends to include more irrelevant areas than VHMT does; while VHMT only focuses on regions richer of edges and textures. It is coherent with quantitative comparisons between UHMT and VHMT in the tables \ref{tab:uhmt}, \ref{tab:vhmt}.

\ifisreport
\else
In contrast to the previous examples, the figure \ref{fig:uhmt_102},\ref{fig:thmt_102},\ref{fig:vhmt_102} shows
\fi
 an opposite case in general outdoor scenes for which (T/V)HMT produces more reasonable saliency maps. While UHMT0 map covers the whole region of sky despite no interesting features, (T/V)HMT0 correctly focuses on interesting but scatter features on the scene. Similarly, (V/T)HMT does extract more meaningful features than UHMT does 
 \ifisreport
\else
 (see the figures \ref{fig:uhmt_102},\ref{fig:thmt_102},\ref{fig:vhmt_102}). 
 \fi
 In addition, the best saliency map THMT0 or THMT5 highlights more discriminant features than AIM/DIS saliency map. Noteworthy, there are significant differences between VHMT1 and THMT1 saliency maps. While THMT1 over-emphasizes edge points and lines between appeared textures such as trees and sky, the VHMT1 is ridged with highlighted regions and does not hint any standing-out areas. Again this discrepancy in performance may be due to nature of orientation selection in each mode (T/V) as THMT favours only features of horizontal, vertical, diagonal directions; while VHMT mode, a rotation invariant scheme, does not have any oriental differentiation.
The third example is chosen such that complex scenes are presented to the saliency methods. 
\ifisreport
\else
In the figures \ref{fig:uhmt_111},\ref{fig:thmt_111},\ref{fig:vhmt_111},
\fi
there are several fruits on the shelf; it is considerably complicated due to richness of edges, textures, as well as colour. In general, all (U/T/V)HMT-MDIS and AIM/DIS only partially succeed in detecting saliency regions from these images since none of them successfully highlight the fruit with distinguished colour on the shelf (the fruit inside a red circle, 
\ifisreport
\else
figures \ref{fig:uhmt_111},\ref{fig:thmt_111},\ref{fig:vhmt_111})
\fi
. Though most MDISs for variety of scale levels, do not explicitly detect that fruit, UHMT3 and UHMT4 salency maps are able to highlight the location of that fruit ( see UHMT3 an UHMT4 saliency maps, the 
\ifisreport
\else
figures \ref{fig:uhmt_111},\ref{fig:thmt_111},\ref{fig:vhmt_111}
\fi
). The sample matches with the fact that UHMT4 data in the table \ref{tab:uhmt} has extremely good performance in all evaluation schemes. Surprisingly, there are some cases when appropriate choices of scales and parameters of predefined HMT models can over-perform all trained HMT models. The interesting examples of the 
\ifisreport
\else
figures \ref{fig:uhmt_111},\ref{fig:thmt_111},\ref{fig:vhmt_111} open
\fi 
 another research direction about how HMT model can be learned optimally; however, it is the question of another research paper.
\ifiselsarticle
\section{Conclusion}
\fi
\ifissvjour
\section{Conclusion}
\fi
\label{sec:conclusion_journal2_2013}
In conclusion, the multiple discriminant saliency (MDIS), a multi-scale extension of DIS \cite{gao2007} under dyadic scale framework, has strong theoretical foundation as it is quantified by information theory and adapted to multiple dyadic-scale structures. The performance of MDIS against AIM and pseudo-DIS is evaluated on a standard database with well-established numerical tools; furthermore, simulation data prove competitiveness of MDIS over AIM and pseudo-DIS in both accuracy and speed. However, MDIS fails to capture salient regions in a few complex scenes; therefore, the next research step is improving MDIS accuracy in such cases. In addition, implementation of MDIS algorithm in embedded systems is also considered as a possible research direction.

\ifissvjour
\bibliographystyle{spmpsci}
\bibliography{../BibRefs/refs}
\fi
\ifiselsarticle
\bibliographystyle{elsarticle-num}
\bibliography{../BibRefs/refs}
\fi

\end{document}